\pdfoutput=1

\documentclass[11pt]{article}

\usepackage{acl}

\usepackage{times}
\usepackage{ragged2e}
\usepackage{latexsym}
\usepackage{lscape}
\usepackage{multirow}
\usepackage[T1]{fontenc}

\usepackage[utf8]{inputenc}

\usepackage{microtype}
\usepackage{adjustbox}
\usepackage{xspace}
\usepackage{amsmath}
\usepackage{array}
\usepackage{float}

\newcommand{\Drug}{\texttt{DRUG}\xspace}
\newcommand{\ADE}{\texttt{ADE}\xspace}

\newcommand{\Frequency}{\texttt{FREQUENCY}\xspace}
\newcommand{\Dosage}{\texttt{DOSAGE}\xspace}
\newcommand{\Duration}{\texttt{DURATION}\xspace}
\newcommand{\Strength}{\texttt{STRENGTH}\xspace}
\renewcommand{\Form}{\texttt{FORM}\xspace}
\newcommand{\awesome}{\texttt{awesome-align}\xspace}
\newcommand{\fastalign}{\texttt{fast\_align}\xspace}
\newcommand{\ttrain}{\texttt{translate-train}\xspace}
\newcommand{\ttest}{\texttt{translate-test}\xspace}
\newcommand{\mednerf}{MedNERF\xspace}

\title{Multilingual Clinical NER: Translation or Cross-lingual Transfer?}

\author{
Xavier Fontaine$^{1}$\thanks{~~Equal contribution.}~~~
Félix Gaschi$^{1,2}$\footnotemark[1]~~~
Parisa Rastin$^{2}$~~~
Yannick Toussaint$^{2}$\\
 \textsuperscript{\rm 1}SAS Posos, France~~~  \textsuperscript{\rm 2}LORIA, France \\
 \texttt{\{xavier,felix\}@posos.fr}\\
  \texttt{\{felix.gaschi,parisa.rastin,yannick.toussaint\}@loria.fr} \\
}

\begin{document}
\maketitle
\begin{abstract}
Natural language tasks like Named Entity Recognition (NER) in the clinical domain on non-English texts can be very time-consuming and expensive due to the lack of annotated data.
Cross-lingual transfer (CLT) is a way to circumvent this issue thanks to the ability of multilingual large language models to be fine-tuned on a specific task in one language and to provide high accuracy for the same task in another language.
However, other methods leveraging translation models can be used to perform NER without annotated data in the target language, by either translating the training set or test set.
This paper compares cross-lingual transfer with these two alternative methods, to perform clinical NER in French and in German without any training data in those languages.
To this end, we release \mednerf a medical NER test set extracted from French drug prescriptions and annotated with the same guidelines as an English dataset.
Through extensive experiments on this dataset and on a German medical dataset~\citep{gernermed}, we show that translation-based methods can achieve similar performance to CLT but require more care in their design. And while they can take advantage of monolingual clinical language models, those do not guarantee better results than large general-purpose multilingual models, whether with cross-lingual transfer or translation.
\end{abstract}

\section{Introduction}

In recent years, pre-trained language models based on the Transformer architecture~\cite{transformers} have demonstrated high performance on many natural language tasks such as Named Entity Recognition (NER), Natural Language Inference or Question-Answering~\citep{bert,roberta}. These models, which are generally pre-trained on general domain data, can be fine-tuned on downstream tasks to achieve state-of-the-art results. Such models can also be adapted to a specific domain such as the legal~\cite{legalbert} or the biomedical fields~\cite{pubmedbert} and can then outperform the general-domain models on domain-specific tasks.

Extracting medical entities from unstructured texts has become an essential tool to structure medical reports, and pre-trained language models have been naturally used to perform this task~\citep{khan2020mt,clinical_transformers}. These models use training data for fine-tuning that come from biomedical NER datasets like NCBI-disease~\cite{ncbi_disease} or n2c2~\cite{n2c2}. However most of these datasets are only available in English and consequently the majority of such medical NER algorithms are developed for the English language, while medical reports or drug prescriptions are rather written in the country's language.

In the clinical domain, non-English datasets to fine-tune a model are even rarer than in the general domain. Even large domain-specific unlabeled corpora are mostly found in English. For example, the biomedical scientific literature is mostly written and available in English.
Moreover, gathering medical texts and annotating them using expert knowledge is very expensive for low-resource languages and more generally for any non-English language. This restricts the development of medical NER models for non-English languages.

Fortunately, Cross-lingual Transfer (CLT) can work around the absence of training data in the target language, by making use of language models pre-trained on multilingual data. CLT consists in applying on a specific task a multilingual large language model (MLLM) fine-tuned in another language for the same task. For example, models like mBERT \cite{devlin-etal-2019-bert} and XLM-R~\cite{conneau-etal-2020-unsupervised} can be fine-tuned on a general-domain NER task in English and provide competitive results when evaluated in other languages on this task \cite{pires-etal-2019-multilingual,wu-dredze-2019-beto}.

Before MLLMs, cross-lingual adaptation was generally tackled using translation methods~\cite{yarowsky-ngai}. The lack of annotated data can be overcome by translating an English training set into the target language and by projecting the annotations with an alignment algorithm. Another approach consists in translating a test set into English to apply an English NER on it and in projecting the predicted labels in the original language.

Recently, translation models based on the Transformer architectures~\citep{fairseq,wmt22} have demonstrated huge improvements over other machine translation algorithms. Such models can therefore be leveraged to perform cross-lingual learning. However, comparing CLT with translation-based methods has attracted little attention and only comparisons in the general domain~\cite{everything} has been done.

Our proposed contribution is three-fold: (1) we perform extensive experiments with CLT and translation-based methods on a medical NER task in French and in German without any training data in those languages, (2) we release \mednerf, a French clinical NER dataset based on drug prescriptions, which serves as a test set for our experiments in French, (3) we demonstrate that CLT and translation provide comparable results for clinical NER, that the choice of the technique depends on the model's size and that using domain-specific models does not necessarily improve the results over using multilingual models.

\section{Related Work}

\paragraph{Multilingual large language models (MLLMs).} MLLMs are the logical multilingual extension of Large Language Models, trained on multilingual corpora. There exist several versions of MLLMs among which the most popular are mBERT~\cite{bert}, which is BERT multilingual version pre-trained on Wikipedia in a hundred different languages instead on only English Wikipedia (and BookCorpus), and XLM-R~\cite{conneau-etal-2020-unsupervised} which is a multilingual Transformer encoder using the same pre-training principles as RoBERTa~\cite{roberta}, trained in 100 languages on a larger and more diverse corpus than Wikipedia.

XLM-R, mBERT and its distilled version distilmBERT \cite{distilbert} are not trained on any parallel data and show nevertheless strong CLT abilities, for example on NER tasks, even between languages which use different sets of characters \cite{pires-etal-2019-multilingual,wu-dredze-2019-beto}.

\paragraph{Translation-based cross-lingual learning.} Cross-lingual learning can also be achieved by using a translation model for either training an algorithm on translated data or using a NER model on translated texts at inference. These techniques~\cite{yarowsky-ngai,yarowsky-etal} make use of translation models and alignment algorithms and have been compared with CLT on general domain tasks by~\citet{everything} who have shown that using translated and aligned training data improved over zero-shot learning for tasks and languages with weak CLT performances. However this comparison has not been done for domain-specific tasks that often require specific language models like in the medical field. These techniques seem nevertheless to provide good results since they have been leveraged to propose the GERNERMED~\cite{gernermed} and GERNERMED++~\cite{gernermed_pp} models which are medical German NER trained on a German automatic translation of the English dataset n2c2~\cite{n2c2}. 

Performing cross-lingual adaptation in a domain-specific setting raises other questions: such as whether domain-specific language models could be leveraged in translation-based methods, or if translation and alignment models fine-tuned on in-domain data can benefit those methods. To the best of our knowledge, those questions have not been addressed in the literature. 

\paragraph{Neural machine translation models.} Current state-of-the-art machine translation algorithms are based on the Transformer architecture~\cite{wmt22} and can be either encoder-decoder models~\cite{helsinki,fair} or decoder-only models~\cite{decoder_translation}. The quality of the translation can be assessed by different metrics such as the traditional BLEU score~\cite{bleu}, which is a statistical algorithm based on matching n-grams between a proposed translation and a reference one. It has been used for years but new scoring algorithms such as COMET~\cite{comet} which leverage large multilingual models seem to provide more accurate evaluations.

\paragraph{Word alignment algorithms.} Word alignment algorithms are designed to provide a mapping between the words of a sentence and those of its translation. They originally relied on statistical features like the \fastalign algorithm~\citep{fastalign} and have been outperformed by models using contextualized embeddings~\citep{simalign,awesome}.

\paragraph{Evaluation corpora for CLT.}
The CLT abilities of MLLMs can be assessed on several tasks in many languages thanks to multilingual benchmarks like XTREME~\cite{xtreme} which cannot be used to evaluate medical NER models since they contain only general-domain tasks.
Despite the existence of non-English medical NER datasets like QUAERO in French~\cite{quaero} or GGPONC in German~\cite{ggponc}, CLT cannot be evaluated on these datasets as there is no English counterpart annotated with the same guidelines, which is a pre-requisite for CLT evaluation.
In order to tackle this issue, \citet{gernermed_pp} have introduced a small test dataset of 30 German medical sentences from Electronic Health Records (EHR) annotated in the same way as the n2c2 dataset and have used it to assess the performances of their GERNERMED++ model. Following their path we propose to release a medical NER dataset in French based on drug prescriptions.


\section{Method and models}
\label{sec:models}

We now describe the different methods we compare to perform NER without any annotations in the target language: cross-lingual transfer and two translation-based methods, where either the train set or the test set is translated. All of them only use the English annotations from the n2c2 dataset (Track 2, Adverse Drug Events and Medication Extraction)~\cite{n2c2}, which is an English dataset of medical entities extracted from EHR.

\subsection{Cross-lingual transfer (CLT)}

The most intuitive way to perform NER without annotations in the target language is CLT which has shown impressive cross-lingual performances~\cite{wu-dredze-2019-beto}. In our setting, we fine-tuned a multilingual large language model to perform NER on the English n2c2 dataset and evaluate on French and German test sets. In our experiments, XLM-R Base is preferred over mBERT as it has the same number of layers but outperforms it on multilingual benchmarks. XLM-R Large is also used to evaluate the impact of model size and distilmBERT, a smaller MLLM obtained by distillation of mBERT, is used to give insights about what is possible with less resources.

With CLT, these models will only see English NER labels during training and will be evaluated on a German and a French NER test set.

\subsection{\ttrain}

The \ttrain approach consists in constructing a translated version of the n2c2 dataset and in training a NER algorithm in French or German on the translated dataset.

The creation of the synthetic translated dataset is done in two steps. First the whole dataset is translated to the target language using a machine translation algorithm~\citep{helsinki,fair}. Then the labels must be aligned, which means identifying in the translated sentences the spans of text corresponding to the original English annotations. This task is tackled using either a statistical model \fastalign~\cite{fastalign} or the neural algorithm \awesome~\cite{awesome}. These alignment tools are applied to the original English sentence and its translation. They provide a mapping between the words of both sentences which is used to transfer the English annotations to the translated sentences. There are cases, namely in German, where the order of the words in a sentence differ from English, when an English annotation corresponds to several disjoint groups of words in the target language. For example, the sentence "She received an additional three units PRBC overnight"\footnote{PRBC stands for "Packed Red Blood Cells".} is translated to "Sie erhielt über Nacht drei weitere PRBC-Einheiten" and the entity "three units" which is translated by "drei Einheiten" is split into two disjoint words. In those cases, all parts of the split entities have been labeled as the original entity.

This method is similar to the one proposed by~\citet{gernermed_pp} and we improve over it by fine-tuning the translation and alignment models on a corpus of parallel medical texts. The influence of using fine-tuned models for translation and alignment is studied in Section~\ref{sec:influence_choice}.

\subsection{\ttest}

The \ttest method consists in translating the data into English at inference time and in applying an English NER model on it. The labels obtained with the NER models can be used to recover the entities in the original text with an alignment algorithm. The major drawback of this method is that it requires to translate the text at inference time while the translation in the \ttrain method occurs only once during training.

\section{Evaluation data}

While the model is trained on the English dataset n2c2 or a translation of it, it must be evaluated on French and German data, annotated similarly to n2c2 to assess its cross-lingual adaptation abilities.

\subsection{The \mednerf dataset}

We release \mednerf\footnote{The dataset is available at \url{https://huggingface.co/datasets/Posos/MedNERF}.}, a Medical NER dataset in the French language. It has been built using a sample of French medical prescriptions annotated with the same guidelines as the n2c2 dataset.

Sentences containing dosage instructions were obtained from a private set of scanned typewritten drug prescriptions. After anonymization of the drug prescriptions we used a state-of-the-art Optical Character Recognition (OCR) software\footnote{\url{https://cloud.google.com/vision}}.
We then discarded low quality sentences from the output of the OCR and we manually identified the sentences containing dosage instructions. For the purpose of this paper only 100 sentences have been randomly sampled and made public through the \mednerf dataset, which is intended to be a test and not a training dataset.

The annotations of the medical sentences use the n2c2 labels \Drug, \Strength, \Frequency, \Duration, \Dosage and \Form. We did not use the \ADE (Adverse Drug Event) label since it is very rare that such entities are present in drug prescriptions. We also discarded the \texttt{ROUTE} and \texttt{REASON} labels as in~\citep{gernermed_pp} because of either their ambiguous definition or the lack of diversity of the matching samples. A total of 406 entities were annotated in 100 sentences (cf. Table \ref{tab:labels_distribution})\footnote{Randomly sampled examples in Appendix \ref{app:examples}.}.

\begin{table}[h!]
    \centering
    \begin{tabular}{c|c}
        \hline
        NER Tag & Count \\
        \hline
         \Drug & 67 \\
         \Strength & 51 \\
         \Frequency & 76 \\
         \Duration & 43 \\
         \Dosage & 76 \\
         \Form & 93 \\
         \hline
         Total & 406 \\
         \hline         
    \end{tabular}
    \caption{Distribution of labels in \mednerf.}
    \label{tab:labels_distribution}
\end{table}

\subsection{The GERNERMED test dataset}

The evaluation of the different cross-lingual adaptation techniques in German is done using the GERNERMED test set released by~\citet{gernermed_pp}, which consists of 30 sentences from physicians annotated with the same guidelines as n2c2. Table~\ref{tab:datasets} provides statistics about the different datasets used in this paper.

\begin{table}[h!]
    \centering
    \begin{tabular}{c|c|c|c}
        \hline
        dataset & lang. &  sent. & entities \\
        \hline 
        n2c2 & en & 16,656 & 65,495 \\
        GERNERMED-test & de & 30 & 119 \\
        \mednerf & fr & 100 & 406 \\
        \hline
    \end{tabular}
    \caption{Statistics about the datasets.}
    \label{tab:datasets}
\end{table}

\section{Pre-selecting translation and alignment}

The translation-based methods require a translation and an alignment models. We present in this section how we fine-tuned translation and alignment algorithms and how we chose which algorithms to use in our experiments. The choice of these algorithms can be seen as an hyper-parameter and for fair comparison with CLT, the selection should not be based on downstream cross-lingual abilities as this would mean cross-lingual supervision.


\subsection{Translation models}
\label{sec:translation_models}

We perform the automated translation of the n2c2 dataset from English to French and German with the following transformer-based machine translation algorithms: Opus-MT~\cite{helsinki} and FAIR~\cite{fair} which we fine-tuned on a corpus of bilingual medical texts proposed in the BioWMT19 challenge~\footnote{The links of the datasets of the BioWMT19 challenge are available on its page \url{https://www.statmt.org/wmt19/biomedical-translation-task.html}}~\cite{biowmt19}. We used the UFAL dataset, which is a collection of medical and general domain parallel corpora in 8 languages paired with English, and Medline which is a dataset containing the titles and abstracts of scientific publications from Pubmed in English and a foreign language.

Since the UFAL dataset is orders of magnitude larger than Medline, we downsampled it to have equal proportions of sentences coming from Medline, from the medical part of UFAL and from UFAL general data. This resulted in approximately 90k sentences for the German translation models and 164k sentences for translation into French.

We fine-tuned the Opus-MT model~\cite{helsinki} for translation to French and German and the FAIR model~\cite{fair} only for translation to German as no version of it is available in French.

The quality of the different translation models is measured on the Medline test set of the BioWMT~19 challenge and on the Khresmoi dataset~\cite{khresmoi}. Results are presented in Tables~\ref{tab:translation_en_de} and~\ref{tab:translation_en_fr}.

\begin{table}[h!]
    \centering
    \adjustbox{max width=\linewidth}{
    \begin{tabular}{c|cc|cc}
        \hline
         \multirow{2}*{model} & \multicolumn{2}{|c|}{BioWMT19} & \multicolumn{2}{c}{Khresmoi}  \\
          & BLEU & COMET & BLEU & COMET \\
        \hline
         FAIR & \underline{32.8} & 0.628 & \textbf{33.7} & \textbf{0.667} \\
         + ft & \textbf{34.2} & \textbf{0.734} & \underline{32.4} & \underline{0.666} \\
         \hline
         Opus & 32.2 & 0.651 & \underline{32.4} & 0.608 \\
         + ft & 32.5 & \underline{0.700} & 30.5 & 0.619 \\
        \hline
    \end{tabular}
    }
    \caption{Evaluation of the translation models from English to German. Best model in bold and second underlined. ft for finetuned.}
    \label{tab:translation_en_de}
\end{table}

\begin{table}[h!]
    \centering
    \adjustbox{max width=\linewidth}{
    \begin{tabular}{c|cc|cc}
        \hline
         \multirow{2}*{model} & \multicolumn{2}{c|}{BioWMT19} & \multicolumn{2}{c}{Khresmoi}  \\
          & BLEU & COMET & BLEU & COMET \\
        \hline
         Opus & 35.9 & 0.672 & \textbf{48.0} & \textbf{0.791} \\
         + ft & \textbf{36.7} & \textbf{0.786} & 46.5 & \textbf{0.791} \\
        \hline
    \end{tabular}
    }
    \caption{Evaluation of the translation models from English to French.}
    \label{tab:translation_en_fr}
\end{table}

The analysis of these results lead us to choose the FAIR fine-tuned model as the best translation model for English to German translation and the Opus-MT fine-tuned model as the best translation model for English to French translation.

The same pre-selection evaluation was performed for the \ttest approach, with translation models from German and French to English. The models were fine-tuned on the same parallel dataset and similar results led to the same choice of best translation models (cf. Appendix~\ref{sec:ttest-preselection}).

\subsection{Alignment models}
\label{sec:alignment_models}

\fastalign and \awesome are two popular choices for word alignment \cite{everything}. Since \awesome can be fine-tuned on parallel data we use the data used for translation to fine-tune the alignment model.

Choosing the right alignment models for the task can be tricky. While parallel corpora for fine-tuning \awesome might be available in several languages and domains, annotated word alignment on parallel data is more scarce. In our case, annotated word alignment test data is not available in the clinical domain. The best alignment models can thus only be selected based on performance on a general-domain dataset. \awesome pre-trained on general-domain data is preferred in French, and the same model with further fine-tuning on biomedical data is selected for German. 

\begin{table}[h!]
    \centering
    
    \begin{tabular}{c|cc}
        \hline
        model & fr & de \\
        \hline
        FastAlign  &  10.5 & 27.0 \\
        \hline 
        AWESOME from scratch & 5.6 & 17.4 \\
        + ft on clinical & 4.7 & 15.4 \\
        \hline
        AWESOME pre-trained & \textbf{4.1} & 15.2 \\
        + ft on clinical & 4.8 & \textbf{15.0} \\
        \hline
    \end{tabular}
    \caption{Average Error Rate (AER) for various aligners.}
    \label{tab:alignment_models}
\end{table}

Table~\ref{tab:pre_selection} summarizes the choices of the best translation and alignment methods.

\begin{table}[h!]
    \centering
    
    \begin{tabular}{c|cc}
        \hline
        lang & translation & alignment \\
        \hline
        fr & Opus ft & AWESOME \\
        de & FAIR ft & AWESOME pt+ft\\
        \hline
    \end{tabular}
    \caption{Pre-selected translation and alignment models.}
    \label{tab:pre_selection}
\end{table}

\section{Results}

Having selected the best translation and alignment model for each language based on intrinsic evaluation, \ttrain and \ttest approaches can be compared to Cross-lingual Transfer (CLT). The impact of the translation and alignment models can also be analyzed, as well as using the translation fine-tuning data to improve directly the models used in CLT. Since \ttrain can leverage monolingual domain-specific model, we evaluate the outcome of such a strategy. Five different random seeds were used for each model and results presented in this section show the average performance along with the standard deviation (more implementation details in Appendix \ref{annex:implementation_details}).

\subsection{Comparison of the different methods}

For a fair comparison between the three methods detailed in Section~\ref{sec:models}, we use the pre-selected translation and alignment models of Table~\ref{tab:pre_selection} for the \ttrain and \ttest methods. We report the F1-scores of the different methods in Table~\ref{tab:main_comparison}. The translation and alignment models providing the best test scores are also provided for comparison, revealing what can be missed with the pre-selection.

\begin{table}
    \centering
    \begin{tabular}{l|cc}
        \hline
        method  & fr & de \\
        \hline 
        \hline
        \multicolumn{3}{c}{distilmBERT} \\
        \hline
        CLT & 65.9$_{\pm 3.3}$ & 64.6$_{\pm 2.4}$\\
        translate-train select. & 66.5$_{\pm 1.9}$ & \textbf{68.3$_{\pm 1.3}$} \\
        translate-test select. & \textbf{69.2$_{\pm 1.4}$} & \textbf{68.3$_{\pm 1.8}$}\\
        \hline
        \textit{translate-train best} & \textit{\underline{69.2}$_{\pm 1.2}$} & \textit{\underline{69.2}$_{\pm 1.2}$} \\
        \textit{translate-test best} & \textit{\underline{69.7}$_{\pm 1.5}$} & \textit{\underline{68.3}$_{\pm 1.8}$}  \\
       \hline 
       \hline
        \multicolumn{3}{c}{XLM-R Base} \\
        \hline
        CLT & \textbf{79.1$_{\pm 0.8}$} & 72.2$_{\pm 0.7}$\\
        translate-train select. & 74.6$_{\pm 0.9}$ & \textbf{73.7$_{\pm 0.9}$} \\
        translate-test select. & 74.2$_{\pm 1.6}$ & 72.7$_{\pm 0.8}$ \\
        \hline
        \textit{translate-train best} & \textit{78.6$_{\pm 0.5}$} & \textit{\underline{74.8}$_{\pm 1.0}$} \\
        \textit{translate-test best} & \textit{74.4$_{\pm 1.3}$} & \textit{72.7$_{\pm 0.8}$} \\
       \hline 
       \hline
        \multicolumn{3}{c}{XLM-R Large} \\
        \hline
        CLT & \textbf{77.9$_{\pm 1.7}$} & \textbf{78.5$_{\pm 0.4}$}\\
        translate-train select. & 76.5$_{\pm 0.7}$ & 77.4$_{\pm 1.3}$ \\
        translate-test select. & 75.3$_{\pm 0.9}$ & 76.1$_{\pm 2.8}$ \\
        \hline
        \textit{translate-train best} & \textit{\underline{78.0}$_{\pm 0.5}$} & \textit{\underline{79.4}$_{\pm 1.3}$} \\
        \textit{translate-test best} & \textit{75.3$_{\pm 0.9}$} & \textit{76.1$_{\pm 2.8}$} \\
        \hline
    \end{tabular}
    \caption{Comparing the three methods with pre-selected translation and alignment models (select.). Best performing pairs are provided for comparison and are underlined when better than CLT.}
    \label{tab:main_comparison}
\end{table}

CLT with a sufficiently large MLLM provides the best results. When compared with translation-based methods with pre-selected translation and alignment models, CLT with XLM-R models gives higher scores, except for XLM-R Base in German. 

On the other hand, it seems that using an English NER on a translated version of the test set provides the best results with a small model like distilmBERT. DistilmBERT might be better as a monolingual model than a multilingual one. In the same vein, XLM-R Base struggles in German, while its large version does not. Small language models underperform in CLT and their generalization ability is not sufficient compared to translation-based methods. A first take-away is consequently that translation-based methods should be favored with small language models.

The \ttest method is consistently outperformed by \ttrain for large models. Even using a specific biomedical model like~\citep{pubmedbert} for \ttest does not improve the results (results in Appendix~\ref{app:full_results}). Indeed translation and alignment errors only harm the training set of \ttrain, which does not prevent a large model from generalizing despite some errors in the training data, while errors of translation or alignment in \ttest are directly reflected in the test score. In the rest of this analysis we will consequently compare CLT only with \ttrain.

Providing a large-enough MLLM, CLT outperforms pre-selected \ttrain and \ttest. However, choosing the translation and the alignment model beforehand does not lead to the best results. To the exception of XLM-R Base in French, there always exists a pair of translation and alignment models that leads to a better score for the \ttrain method over CLT. This agrees with~\citet{everything} and encourages practitioners to explore different methods to perform cross-lingual adaptation.

\subsection{Influence of the translation model}
\label{sec:influence_choice}

The choice of the translation and alignment models can have an important impact on the final NER performances as shown on Table~\ref{tab:main_comparison}. This section studies their impact in details.
A German NER model was trained using the Opus model instead of the FAIR model for translating the training set into German. Using a worse model (see Table~\ref{tab:translation_en_de}) for translation leads to lower NER scores as shown in Table~\ref{tab:german_helsinki}: the NER model based on the FAIR translation beats by more than 2 points the one using the Opus translation, whatever aligner is used.

While choosing between different base translation models (like Opus or FAIR) based on their translation scores on in-domain data seems to provide the best results, deciding between the fine-tuned version of a translation model and the base one by comparing the BLEU or COMET scores on biomedical data does not guarantee the best downstream F1 score as Table~\ref{tab:ft_vs_base} shows. The translation model was fine-tuned on biomedical data, which improved intrinsic results on the BioWMT19 translation dataset. But this dataset belongs to a specific biomedical sub-domain (PubMed abstracts), and fine-tuning might not improve translation for the clinical sub-domain of the NER dataset.

The takeaway is that, while a small gain in translation accuracy (obtained with further fine-tuning) might not necessarily improve the result of the \ttrain approach, a completely different model (like FAIR with respect to Opus) has more chance to improve cross-lingual adaptation. 

\begin{table}
    \centering
    
    \begin{tabular}{p{3cm}|p{1.7cm}p{1.7cm}}
    \hline
         aligner & Opus f1 & FAIR f1 \\
\hline
FastAlign &  70.9$_{\pm 1.8}$ & \textbf{72.8$_{\pm 1.6}$} \\
AWESOME &  72.2$_{\pm 1.7}$ & \textbf{73.1$_{\pm 1.3}$} \\
AWESOME ft & 71.1$_{\pm 1.2}$ & \textbf{74.1$_{\pm 1.1}$}\\
AWESOME pt+ft & 71.2$_{\pm 1.1}$ & \textbf{74.1$_{\pm 1.3}$} \\
\hline
    \end{tabular}
    \caption{\ttrain in German with XLM-R Base using either fine-tuned or base Opus model.}
    \label{tab:german_helsinki}
\end{table}

\begin{table}[]
    \centering
    
    \begin{tabular}{p{3cm}|p{1.7cm}p{1.7cm}}
    \hline
         aligner & base & fine-tuned \\
\hline
FastAlign &  78.2$_{\pm 0.8}$ & \textbf{78.6$_{\pm 0.5}$} \\
AWESOME &  \textbf{76.4$_{\pm 2.0}$} & 74.2$_{\pm 0.9}$ \\
AWESOME ft & \textbf{74.6$_{\pm 1.6}$} & 74.5$_{\pm 1.6}$\\
AWESOME pt+ft & 75.8$_{\pm 1.7}$ & \textbf{76.3$_{\pm 1.0}$} \\
\hline
    \end{tabular}
    \caption{\ttrain in French with XLM-R Base using either fine-tuned or base Opus model.}
    \label{tab:ft_vs_base}
\end{table}

\subsection{Influence of the alignment model}

While choosing a translation model based solely on intrinsic performance should not harm downstream cross-lingual adaptation performances, the choice of the alignment model seems more tricky. Based on intrinsic performances like Error Rate on annotated alignment (Table~\ref{tab:alignment_models}), \awesome seems to be the right aligner for the task. However, while it provides better downstream results than \fastalign in German (Table~\ref{tab:german_helsinki}), it does not hold for French (Table~\ref{tab:ft_vs_base}).

\begin{table}[]
    \centering
    
    \begin{tabular}{c|ccc|c}
        \hline
        aligner & Freq. & Strength & Drug & f1 \\
        \hline
        FastAlign & \textbf{72.0} & 89.7 & \textbf{83.2} & \textbf{78.6} \\
        AWESOME & 50.2 & \textbf{92.5} & 82.2 & 74.2 \\
        \hline
    \end{tabular}
    \caption{Comparison of \fastalign and \awesome (pre-trained only) for three different entity types (F1-score), for \ttrain with XLM-R Base on \mednerf with Opus fine-tuned. (Full results in Appendix~\ref{app:full_results}).}
    \label{tab:by_class}
\end{table}

Table~\ref{tab:by_class} shows that using different aligners leads to different levels of accuracy according to the types of entity we want to retrieve. While the global F1 score suggests that \fastalign is better suited for the cross-lingual adaptation, looking at the detailed results for each entity type shows that the gap is mainly due to the \Frequency class on which \awesome performs poorly. But this is not the case on other classes.

\begin{table}[]
    \centering
    \adjustbox{max width=\linewidth}{ 
    \begin{tabular}{p{1.7cm}|p{3.8cm}p{2.8cm}}
        \hline
        original & FastAlign & AWESOME \\
        \hline
        in morning and night & le matin et la nuit et 240 mg & \textbf{le matin et la nuit} \\
        \hline
        daily & \textbf{par jour} & jour \\
        \hline
        once a day & \textbf{une fois par jour} & fois par jour \\
        \hline
        at bedtime & \textbf{au moment du coucher} & au \textbf{//} coucher \\
        \hline
    \end{tabular}
    }
    \caption{Examples of frequencies transformed with translation and alignment. Bold indicates the right annotation and \textbf{//} indicates that the entity has been split.}
    \label{tab:frequency_examples}
\end{table}

\Frequency entities are usually more verbose than drugs or dosages. Table~\ref{tab:frequency_examples} shows that \fastalign make obvious errors like aligning "240 mg" to "in morning and night", but \awesome can miss the preposition when aligning "daily" with "jour" instead of "par jour", leading eventually to a consequent score drop. 

The choice of the alignment model must thus be made more carefully than the translation one. Intrinsic performances of alignment models are not sufficient information. Some additional post-processing might be needed, as in \citet{everything}, where \awesome gives better results, but entities that are split by the aligner like "au moment du coucher" in Table~\ref{tab:frequency_examples} are merged by including all words in between. This would work in that particular case, but could cause problems in others, particularly for languages where the word order is different.

\subsection{Using parallel data to realign models}

With the right translation and alignment model, it seems that CLT can be outperformed by the \ttrain method. However the latter relies on additional resources: a translation and an alignment models, trained on parallel data. This parallel data could also be used to re-align the representations of the multilingual models used in CLT.

To improve a multilingual language model with parallel data, it is trained for a contrastive alignment objective following \citet{wu-dredze-2020-explicit}. Words aligned with \awesome are trained to have more similar representations than random in-batch pairs of words (details in Appendix \ref{annex:realignment}). After this realignment step, CLT can be applied.

\begin{table}[h!]
    \centering
    \begin{tabular}{c|cc}
        \hline
        model & fr & de  \\
        \hline
        distilmBERT & 65.9$_{\pm 3.3}$ & 64.6$_{\pm 2.4}$ \\
        + realign  & \underline{66.4}$_{\pm 1.4}$ & \underline{67.9}$_{\pm 1.5}$ \\
        translate-train best & \textbf{69.2$_{\pm 1.2}$} & \textbf{69.2$_{\pm 1.2}$} \\
        \hline
        XLM-R Base & \textbf{79.1$_{\pm 0.8}$} & 72.2$_{\pm 0.7}$ \\
        + realign & 76.7$_{\pm 0.7}$ & \textbf{\underline{75.8}$_{\pm 1.3}$}\\
        translate-train best & 78.6$_{\pm 0.5}$ & 74.8$_{\pm 1.0}$ \\
        \hline
        XLM-R Large & 77.9$_{\pm 1.7}$ & 78.5$_{\pm 0.4}$\\
        + realign & \textbf{\underline{78.8}$_{\pm 1.6}$} & 78.3$_{\pm 1.6}$\\
        translate-train best & 78.0$_{\pm 0.5}$ & \textbf{79.4$_{\pm 1.3}$} \\
        \hline
    \end{tabular}
    \caption{F1 scores for CLT from scratch and CLT with realignment. Best F1-score in bold. Results underlined show improvement of realignment over CLT.}
    \label{tab:realign_results}
\end{table}

Results in Table~\ref{tab:realign_results} show that while realignment does not systematically provide improvement over CLT as observed by \citet{wu-dredze-2020-explicit}, it does significantly boost results in some cases, allowing to outperform the best \ttrain baseline in German for XLM-R Base and in French for XLM-R Large. This, yet again, encourages practitioners to explore different methods, including realignment to perform cross-lingual adaptation. 

\subsection{Using domain-specific language models}

We evaluate now the relevance of using language-specific models like CamemBERT~\cite{camembert} or GottBERT~\cite{gottbert} on the translated version of the training dataset or language and domain-specific models like DrBERT~\cite{drbert} or medBERT.de~\cite{medbert} which are BERT models fine-tuned on medical corpora in respectively French and German. We report in Table~\ref{tab:domain_french_results} and~\ref{tab:domain_german_results} the results of the \ttrain method for the best translation/alignment algorithms pair and for the pre-selected one, compared to using XLM-R Base.

\begin{table}[h!]
    \centering
    \begin{tabular}{p{3cm}|cc}
        \hline
        model & pre-selected & best \\
        \hline
        CamemBERT Base & 73.5$_{\pm 1.5}$ & 76.7$_{\pm 0.9}$ \\
        DrBERT 7GB & 70.7$_{\pm 1.3}$ & 73.5$_{\pm 1.4}$ \\
        DrBERT Pubmed & \textbf{76.1}$_{\pm 1.3}$ & \textbf{78.8}$_{\pm 1.4}$  \\
        XLM-R Base & 74.6$_{\pm 1.9}$ & 78.6$_{\pm 0.5}$ \\
        \hline
    \end{tabular}
    \caption{Comparison of domain and language specific models for translate-train in French.}
    \label{tab:domain_french_results}
\end{table}

\begin{table}[h!]
    \centering
    \begin{tabular}{p{3cm}|cc}
    \hline
        model & pre-selected & best \\
        \hline
        GottBERT & \textbf{75.5}$_{\pm 1.4}$ & \textbf{76.6}$_{\pm 0.8}$  \\
        medBERT & 72.7$_{\pm 0.5}$ & 75.0$_{\pm 1.6}$\\
        XLM-R Base & 73.7$_{\pm 0.9}$ & 74.8$_{\pm 1.0}$\\
        \hline
    \end{tabular}
    \caption{Comparison of domain and language specific models for translate-train in German.}
    \label{tab:domain_german_results}
\end{table}

The \ttrain approach allows to rely on models that are specific to the language and domain of the target evaluation. However, Table~\ref{tab:domain_french_results} and~\ref{tab:domain_german_results} show that their use does not always bring significant improvement over XLM-R Base.
The performances of these models can be explained by the quantity of training data used. XLM-R models are indeed trained on 2.5 TB data while DrBERT and medBERT.de use less than 10GB data, which can explain their low score. Besides, the language-specific models CamemBERT and GottBERT are trained with more data (138 GB and 145 GB) and achieve better performances, even beating XLM-R in German. 
Finally, it must be noted that the best \ttrain model in French, DrBERT Pubmed, is actually pre-trained on the English PubMed dataset and then on French clinical texts, which suggests that multilingual models should be preferred, even with a translation-based cross-lingual adaptation.

\subsection{Computing times}

To conclude the analysis of the different cross-lingual adaptation methods studied in this paper we finally compare their computing times. Table~\ref{tab:compute_time} gathers the training and inference times of the three methods using the XLM-R base model and the \awesome alignment model in French.

\begin{table}[h!]
    \centering
    \begin{tabular}{c|cc}
        \hline
         \multirow{2}*{method} & training time & inference time \\
         & (total) & (per sample) \\
        \hline
        CLT & 1.2h & 0.04s  \\
        translate-train & 2.7h & 0.04s \\
        translate-test & 1.2h & 0.32s  \\
        \hline
    \end{tabular}
    \caption{Training and inference times for the different methods, with the XLM-R Base model in French, on a single GPU.}
    \label{tab:compute_time}
\end{table}

This comparison shows a longer training time for the \ttrain method, which is due to the translation of the whole training set before the training of the NER model. On a single GPU with 8GB of RAM this translation step is even longer than the NER training. However, once training is done the \ttrain method has the same inference time as the CLT method, while the \ttest method now suffers from the need of translation at inference time.

\section{Conclusion}

This paper shows that cross-lingual transfer with general-domain MLLMs is efficient for a domain-specific task like clinical NER, giving comparable results with translating the training set. But CLT has the advantage of working off-the-shelf, while translation-based methods require choosing the translation and alignment models carefully. Selecting these models based on intrinsic domain-specific values, like fine-tuning scores on clinical parallel data, or using a domain-specific language model does not provide significantly better downstream results in the target language. The selection of the alignment model was shown to be particularly crucial, and results of \ttrain could probably be improved by post-processing the alignment. CLT also has a margin of progression as realigning the representations of MLLMs can increase the results dramatically in some cases.

It is also worth noting that training on translated data provide better results than translating at inference time. The \ttest approach should then be used only when large multilingual models cannot be used.
While training on translated data allows to leverage domain-specific monolingual language models, those latter models can give better results over multilingual models like XLM-R only if pre-trained with sufficient data.

Pre-training a MLLM with only clinical data is a good lead for further improvements in clinical cross-lingual transfer. While the results show that using a domain-specific monolingual model in \ttrain or \ttest is not on par with general-purpose multilingual models, they also show that the French clinical model DrBERT provides the best results for \ttrain when it uses the English biomedical model PubmedBERT as initialization.

We finally advocate for the release of more non-English clinical datasets annotated with similar guidelines as English (or other) ones. Even a relatively small dataset like \mednerf or the GERNERMED test set are crucial to evaluate cross-lingual adaptation in the clinical domain.

\section*{Limitations}

This paper is limited to the study of clinical NER models using an encoder-only architecture. The use of generative models with a zero-shot learning approach~\citep{clinicalchatgpt} is another promising approach for low-resource languages that could be compared with CLT and translation-based approaches in a future work. However such methods require a careful prompt selection strategy and cannot be directly compared to supervised models.

This paper is also limited to cross-lingual transfer to French and German. Ideally, this work could have included experiments with other target languages and also other source languages than English, as \citet{everything} do in their general-domain comparison of strategies for cross-lingual transfer. However evaluation datasets are lacking for that purpose in the clinical domain. Similarly, more general conclusions about cross-lingual adaptation methods in the clinical domains could be drawn with further studies on various clinical NLP tasks such as relation extraction or sentence classification. However, the lack of evaluation datasets in the clinical domain prevented us from  extending the experiments to such other clinical NLP tasks. Finally, the authors assume that the findings will be task-specific and encourage practitioners to explore all methods when facing a new NLP task.

The authors also want to point out that \mednerf is drawn from drug prescriptions while the n2c2 and GERNERMED datasets use clinical reports.
This domain difference could have made the cross-lingual generalization more challenging, but in practice we found that the different models used were not really affected by the possible domain-shift, showing similar French and German F1 scores. Moreover, when comparing randomly sampled examples from all three datasets, we do not find any critical differences (see Appendix~\ref{app:examples}). Sentences drawn from \mednerf are shorter and less written, but they contain similar annotated entities as the n2c2 sentences, and the n2c2 dataset also contains some short examples that resemble the \mednerf ones.\looseness=-1

\section*{Acknowledgments}

We would like to thank the anonymous reviewers
for their comments which helped enrich the discussion around the results.

Some experiments presented in this paper were carried
out using the Grid’5000 testbed, supported by a scientific interest group hosted by Inria and including
CNRS, RENATER and several Universities as well
as other organizations\footnote{see \url{https://www.grid5000.fr}.}.

\bibliography{anthology,custom}
\bibliographystyle{acl_natbib}

\clearpage
\appendix

\section{Pre-selection of the translation model for \ttest}\label{sec:ttest-preselection}

The \ttest method needs translation algorithms from German and French to English. Similarly to Section~\ref{sec:translation_models} we fine-tuned the Opus-MT and FAIR algorithms on the same medical datasets and obtained the COMET and BLEU scores presented in Tables~\ref{tab:translation_de_en} and~\ref{tab:translation_fr_en}. These scores are used to select the best translation model for the \ttest approach and they lead to the same model choices as for the \ttrain method.

\begin{table}[h!]
    \centering
    \adjustbox{max width=\linewidth}{
    \begin{tabular}{c|cc|cc}
        \hline
         \multirow{2}*{model} & \multicolumn{2}{|c|}{BioWMT19} & \multicolumn{2}{|c}{Khresmoi}  \\
          & BLEU & COMET & BLEU & COMET \\
        \hline
         FAIR & \underline{38.2} & 0.538 & \textbf{47.1} & \textbf{0.764} \\
         + ft & \textbf{38.5} & \textbf{0.675} & \underline{46.8} & \textbf{0.764} \\
         \hline
         Opus & 35.3 & 0.587 & 43.6 & 0.723 \\
         + ft & 38.1 & \underline{0.640} & 44.3 & 0.729 \\
        \hline
    \end{tabular}
    }
    \caption{Evaluation of the translation models from German to English. Best model bold and second underlined. ft for finetuned.}
    \label{tab:translation_de_en}
\end{table}

\begin{table}[h!]
    \centering
    \adjustbox{max width=\linewidth}{
    \begin{tabular}{c|cc|cc}
        \hline
         \multirow{2}*{model} & \multicolumn{2}{|c|}{BioWMT19} & \multicolumn{2}{|c}{Khresmoi}  \\
          & BLEU & COMET & BLEU & COMET \\
        \hline
         Opus & 33.9 & 0.721 & \textbf{48.3} & 0.798 \\
         + ft & \textbf{36.3} & \textbf{0.749} & 48.0 & \textbf{0.799} \\
        \hline
    \end{tabular}
    }
    \caption{Evaluation of the translation models from French to English. Best model bold. ft for finetuned.}
    \label{tab:translation_fr_en}
\end{table}

\section{Training and implementation details}\label{annex:implementation_details}

All models were written in Pytorch using the Huggingface librairies~\citep{huggingface} and were fine-tuned using the AdamW optimizer~\citep{adamw} and a learning rate of $6\cdot 10^{-6}$ with linear decay. We used 4 epochs for the translation models and 8 epochs for the NER models. The NER models were trained on a single GPU (Nvidia GeForce RTX 2070 with 8GB of RAM) for approximately one hour.

\section{Alignment methods used}

\fastalign was applied asymmetrically, by mapping words from source language (English) to target language. Although it might increase alignment score, symmetrization was not used, because it might remove important links during the labels projection step. 

\awesome was used with softmax (instead of the alternative $\alpha$-entmax function) and without the optional consistency optimization objective. For completeness we added the results with the consistency optimization objective (w/ co lines) in the tables of Appendix~\ref{app:full_results} and we observed that they did not improve the NER scores. The base model used is mBERT as in the original paper \cite{awesome}. The pre-trained version of \awesome used is the one provided by the authors, fine-tuned on general-domain parallel data. Throughout the paper, in tables, "AWESOME" designates this latter pre-trained version. "AWESOME ft" is \awesome with the raw mBERT model fine-tuned on the clinical parallel data only and "AWESOME pt+ft" is the pre-trained model, fine-tuned again on the clinical parallel data.

\section{Transformer-base models used}
\label{appendix:model_sizes}

We report in Table~\ref{tab:model_sizes} the number of parameters and the quantity of training data of the different large language models used in this paper.

\begin{table}[h!]
    \centering
    \begin{tabular}{c|ccc}
        \hline
        \multirow{2}*{model} & params & emb. & train \\
        & (M) & (M) & (GB)\\
        \hline
        \multicolumn{4}{l}{Multilingual models}\\
        \hline
        distilmBERT & 135 & 92 & 42 \\
        XLM-R Base & 278 & 192 & 2.5k \\
        XLM-R Large & 560 & 256 & 2.5k \\
        \hline
        \multicolumn{4}{l}{Language-specific models}\\
        \hline
        CamemBERT (fr) & 111 & 25 & 138 \\
        GottBERT (de) & 126 & 40 & 145 \\
        \hline
        \multicolumn{4}{l}{Clinical models}\\
        \hline
        medBERT (de) & 109 & 23 & 10 \\
        DrBERT 7GB (fr) & 111 & 25 & 7.4 \\
        DrBERT PubMed (fr) & 109 & 23 & 28 \\
        \hline
    \end{tabular}
    \caption{Size of the different base models.}
    \label{tab:model_sizes}
\end{table}

Although distilmBERT has more parameters than CamemBERT, it must be noted that it has also more words in its vocabulary, due to its multilingual nature. Hence most of its parameters are embeddings weights that are not necessarily used in our experiments as they might be embeddings of words from other languages. So in our setting, distilmBERT can be considered a smaller model than CamemBERT and GottBERT despite the higher number of parameters.

\section{Realignment method}\label{annex:realignment}

The reader might refer to \citet{wu-dredze-2020-explicit} for the realignment method itself. The representations of the last layer of the pre-trained model to be realigned were used in a contrastive loss where pairs of words aligned with \awesome are encouraged to be more similar than the other possible pairs of words in the batch. The strong alignment objective was used, meaning that pair of same-language words were also used as negative examples for the contrastive loss. The version of \awesome was the one pre-trained on general-domain data, released by \citet{awesome}, with softmax and without the optional consistency optimization objective, the same used by the authors of the realignment method used.

The parallel data used for realignment was the same as for fine-tuning the translation and alignment models (Section \ref{sec:translation_models}). The two datasets (English-German and English-French) were used together to realign a given model, which can then be used either for generalization to French or German. For each base model, five realigned models were obtained for the five random seeds, each of them used in the corresponding fine-tuning by seed. 

The realignment was done for 20,000 steps of batches of size 16, with Adam optimizer, a learning rate of $2\times 10^{-4}$, and with linear warm-up for 10\% of the total steps. This means that the whole dataset was repeated approximately 1.25 times.

\section{Additional results}
\label{app:full_results}


Detailed results are shown in the following tables:

\begin{itemize}
    \item Summary of results for cross-lingual adaptation, with pre-selected and best pairs of translation and alignment models: Table \ref{tab:summary_fr} for French and \ref{tab:summary_de} for German;
    \item CLT and \ttrain with multilingual models: Table \ref{tab:full_french} (fr) and \ref{tab:full_german} (de);
    \item \ttrain with language- and domain-specific models: Table \ref{tab:full_specific_french} and \ref{tab:full_specific_german};
    \item \ttest with multilingual language models: Table \ref{tab:full_test_fr} and \ref{tab:full_test_de};
    \item \ttest with PubmedBERT: Table \ref{tab:pubmed_test_fr} and \ref{tab:pubmed_test_de};
    \item Breakdown of the results class-by-class in French for multilingual models: Table \ref{tab:by_class_details}.
\end{itemize}

\begin{table}[]
    \centering
    \begin{tabular}{c|cc}
        \hline
        model & pre-selected & best \\
        \hline
        \multicolumn{3}{c}{translate-train}\\
        \hline
        distilmBERT & 66.5$_{\pm 1.9}$ & 69.2$_{\pm 1.2}$ \\
        XLM-R Base & 74.2$_{\pm 0.9}$ & 78.6$_{\pm 0.5}$\\
        XLM-R Large & 76.5$_{\pm 0.7}$ & 78.0$_{\pm 0.5}$ \\
        CamemBERT & 73.5$_{\pm 1.5}$ & 76.7$_{\pm 0.9}$ \\
        DrBERT & 70.7$_{\pm 1.3}$ & 73.5$_{\pm 1.4}$ \\
        DrBERT Pubmed & 76.1$_{\pm 1.3}$ & 78.8$_{\pm 1.4}$ \\
        \hline
        \multicolumn{3}{c}{translate-test}\\
        \hline
        distilmBERT & 69.2$_{\pm 1.4}$ & 69.7$_{\pm 1.5}$ \\
        XLM-R Base & 74.2$_{\pm 1.6}$ & 74.4$_{\pm 1.3}$\\
        XLM-R Large & 75.3$_{\pm 0.9}$ & 75.3$_{\pm 0.9}$\\
        PubmedBERT & 73.3$_{\pm 1.3}$ & 73.5$_{\pm 1.2}$ \\
        \hline 
        \multicolumn{3}{c}{CTL$^*$}\\
        \hline
        distilmBERT & 65.9$_{\pm 3.3}$ & 65.9$_{\pm 3.3}$\\
        + realigned & 66.4$_{\pm 1.4}$ & 66.4$_{\pm 1.4}$ \\
        XLM-R Base & \textbf{79.1$_{\pm 0.8}$}& \textbf{79.1$_{\pm 0.8}$}\\
        + realigned & 76.7$_{\pm 0.7}$& 76.7$_{\pm 0.7}$\\
        XLM-R Large & 77.9$_{\pm 1.7}$& 77.9$_{\pm 1.7}$\\
        + realigned & 78.8$_{\pm 1.6}$& 78.8$_{\pm 1.6}$\\
        \hline
    \end{tabular}
    \caption{Summary of results for cross-lingual adaptation to French.}
    {\footnotesize $^*$results are reported twice as there is no pre-selection process}
    \label{tab:summary_fr}
\end{table}

\begin{table}[]
    \centering
    \begin{tabular}{c|cc}
        \hline
        model & pre-selected & best \\
        \hline
        \multicolumn{3}{c}{translate-train}\\
        \hline
        distilmBERT & 68.3$_{\pm 1.3}$ & 69.2$_{\pm 1.2}$ \\
        XLM-R Base & 73.7$_{\pm 0.9}$ & 74.8$_{\pm 1.0}$\\
        XLM-R Large & 77.4$_{\pm 1.3}$ & \textbf{79.4$_{\pm 1.3}$} \\
        GottBERT & 75.5$_{\pm 1.4}$ & 76.6$_{\pm 0.8}$ \\
        MedBERT.de & 72.7$_{\pm 0.5}$ & 75.0$_{\pm 1.6}$ \\
        \hline
        \multicolumn{3}{c}{translate-test}\\
        \hline
        distilmBERT & 68.3$_{\pm 1.8}$ & 68.3$_{\pm 1.8}$ \\
        XLM-R Base & 72.7$_{\pm 0.8}$ & 72.7$_{\pm 0.8}$\\
        XLM-R Large & 76.1$_{\pm 2.8}$ & 76.1$_{\pm 2.8}$\\
        PubmedBERT & 72.6$_{\pm 1.5}$ & 73.3$_{\pm 1.7}$ \\
        \hline 
        \multicolumn{3}{c}{CTL$^*$}\\
        \hline
        distilmBERT & 64.6$_{\pm 2.4}$ & 64.6$_{\pm 2.4}$\\
        + realigned & 67.9$_{\pm 1.5}$ & 67.9$_{\pm 1.5}$\\
        XLM-R Base & 72.2$_{\pm 0.7}$  & 72.2$_{\pm 0.7}$\\
        + realigned & 75.8$_{\pm 1.3}$& 75.8$_{\pm 1.3}$\\
        XLM-R Large & \textbf{78.5$_{\pm 0.4}$}& 78.5$_{\pm 0.4}$\\
        + realigned & 78.3$_{\pm 1.6}$& 78.3$_{\pm 1.6}$\\
        \hline
    \end{tabular}
    \caption{Summary of results for cross-lingual adaptation to German.}
    {\footnotesize $^*$results reported twice as there is no pre-selection process}
    \label{tab:summary_de}
\end{table}

\begin{table*}
\centering
\adjustbox{totalheight=\textheight-2\baselineskip}{
    \begin{tabular}{cc|cccc}
\hline
translation & aligner & precision & recall & micro-f1 & macro-f1\\
\hline
\multicolumn{6}{c}{distilmBERT}\\
\hline
Opus & FastAlign & 67.8$_{\pm 2.1}$ & 68.7$_{\pm 1.5}$ & 68.3$_{\pm 1.6}$ & 70.7$_{\pm 1.4}$\\
Opus & AWESOME w/o co & 67.1$_{\pm 1.0}$ & 68.7$_{\pm 1.3}$ & 67.9$_{\pm 0.3}$ & 70.4$_{\pm 0.3}$\\
Opus & AWESOME w/ co & \textbf{68.2$_{\pm 1.6}$} & 70.1$_{\pm 1.1}$ & \textbf{69.2$_{\pm 1.2}$} & \textbf{71.7$_{\pm 1.1}$}\\
Opus & AWESOME ft w/o co & 65.0$_{\pm 0.8}$ & 67.4$_{\pm 1.8}$ & 66.2$_{\pm 1.0}$ & 68.8$_{\pm 0.8}$\\
Opus & AWESOME ft w/ co & 65.2$_{\pm 1.2}$ & 68.4$_{\pm 1.8}$ & 66.8$_{\pm 1.3}$ & 69.4$_{\pm 1.1}$\\
Opus & AWESOME pt+ft w/o co & 66.5$_{\pm 1.3}$ & 69.1$_{\pm 1.2}$ & 67.8$_{\pm 1.0}$ & 70.2$_{\pm 0.9}$\\
Opus & AWESOME pt+ft w/ co & 66.6$_{\pm 1.2}$ & \textbf{70.2$_{\pm 1.8}$} & 68.3$_{\pm 1.1}$ & 70.8$_{\pm 1.1}$\\
Opus ft & FastAlign & 66.0$_{\pm 1.1}$ & 69.5$_{\pm 0.4}$ & 67.7$_{\pm 0.5}$ & 70.0$_{\pm 0.5}$\\
Opus ft & AWESOME w/o co & 65.2$_{\pm 1.9}$ & 67.8$_{\pm 1.9}$ & 66.5$_{\pm 1.9}$ & 69.1$_{\pm 1.7}$\\
Opus ft & AWESOME w/ co & 65.7$_{\pm 1.2}$ & 68.3$_{\pm 1.9}$ & 66.9$_{\pm 1.4}$ & 69.9$_{\pm 1.2}$\\
Opus ft & AWESOME ft w/o co & 64.6$_{\pm 0.8}$ & 67.4$_{\pm 1.0}$ & 65.9$_{\pm 0.4}$ & 68.6$_{\pm 0.5}$\\
Opus ft & AWESOME ft w/ co & 64.9$_{\pm 0.7}$ & 68.8$_{\pm 1.5}$ & 66.8$_{\pm 1.1}$ & 69.4$_{\pm 1.0}$\\
Opus ft & AWESOME pt+ft w/o co & 64.5$_{\pm 1.1}$ & 68.1$_{\pm 1.4}$ & 66.2$_{\pm 1.0}$ & 69.0$_{\pm 1.0}$\\
Opus ft & AWESOME pt+ft w/ co & 63.5$_{\pm 1.0}$ & 68.3$_{\pm 1.1}$ & 65.8$_{\pm 0.9}$ & 68.7$_{\pm 1.0}$\\
\hline
\multicolumn{2}{c|}{Cross-lingual Transfer} & 64.9$_{\pm 2.5}$ & 66.8$_{\pm 4.1}$ & 65.9$_{\pm 3.3}$ & 68.2$_{\pm 3.2}$\\
\multicolumn{2}{c|}{Cross-lingual Transfer with realignment}  & 68.1$_{\pm 2.2}$ & 64.9$_{\pm 1.0}$ & 66.4$_{\pm 1.4}$ & 67.4$_{\pm 1.6}$\\
\hline
\multicolumn{6}{c}{XLM-R Base}\\
\hline
Opus & FastAlign & 77.4$_{\pm 0.9}$ & 79.1$_{\pm 0.7}$ & 78.2$_{\pm 0.8}$ & 79.8$_{\pm 0.6}$\\
Opus & AWESOME w/o co & 75.4$_{\pm 2.4}$ & 77.3$_{\pm 1.6}$ & 76.4$_{\pm 2.0}$ & 78.6$_{\pm 1.6}$\\
Opus & AWESOME w/ co & 76.0$_{\pm 1.0}$ & 77.6$_{\pm 1.6}$ & 76.8$_{\pm 1.2}$ & 78.8$_{\pm 1.2}$\\
Opus & AWESOME ft w/o co & 73.8$_{\pm 1.7}$ & 75.4$_{\pm 1.6}$ & 74.6$_{\pm 1.6}$ & 76.8$_{\pm 1.4}$\\
Opus & AWESOME ft w/ co & 75.0$_{\pm 0.8}$ & 76.9$_{\pm 0.6}$ & 76.0$_{\pm 0.5}$ & 78.2$_{\pm 0.6}$\\
Opus & AWESOME pt+ft w/o co & 74.8$_{\pm 1.5}$ & 76.8$_{\pm 2.0}$ & 75.8$_{\pm 1.7}$ & 78.0$_{\pm 1.7}$\\
Opus & AWESOME pt+ft w/ co & 75.1$_{\pm 1.3}$ & 76.8$_{\pm 1.6}$ & 76.0$_{\pm 1.4}$ & 78.2$_{\pm 1.2}$\\
Opus ft & FastAlign & 77.9$_{\pm 0.2}$ & 79.4$_{\pm 1.0}$ & 78.6$_{\pm 0.5}$ & 80.3$_{\pm 0.3}$\\
Opus ft & AWESOME w/o co & 72.9$_{\pm 1.4}$ & 75.6$_{\pm 0.8}$ & 74.2$_{\pm 0.9}$ & 76.7$_{\pm 0.7}$\\
Opus ft & AWESOME w/ co & 74.6$_{\pm 1.0}$ & 76.6$_{\pm 0.8}$ & 75.6$_{\pm 0.9}$ & 77.9$_{\pm 0.7}$\\
Opus ft & AWESOME ft w/o co & 73.4$_{\pm 1.7}$ & 75.6$_{\pm 1.6}$ & 74.5$_{\pm 1.6}$ & 77.0$_{\pm 1.4}$\\
Opus ft & AWESOME ft w/ co & 74.2$_{\pm 1.9}$ & 76.7$_{\pm 2.2}$ & 75.5$_{\pm 2.0}$ & 78.0$_{\pm 1.7}$\\
Opus ft & AWESOME pt+ft w/o co & 75.1$_{\pm 1.0}$ & 77.5$_{\pm 1.1}$ & 76.3$_{\pm 1.0}$ & 78.5$_{\pm 0.9}$\\
Opus ft & AWESOME pt+ft w/ co & 75.1$_{\pm 1.9}$ & 77.6$_{\pm 1.7}$ & 76.3$_{\pm 1.8}$ & 78.7$_{\pm 1.5}$\\
\hline
\multicolumn{2}{c|}{Cross-lingual Transfer} & \textbf{78.7$_{\pm 1.8}$} & \textbf{79.6$_{\pm 0.5}$} & \textbf{79.1$_{\pm 0.8}$} & \textbf{80.9$_{\pm 0.9}$}\\
\multicolumn{2}{c|}{Cross-lingual Transfer with realignment}  & 76.9$_{\pm 1.4}$ & 76.6$_{\pm 0.2}$ & 76.7$_{\pm 0.7}$ & 78.9$_{\pm 0.8}$\\
\hline
\multicolumn{6}{c}{XLM-R Large}\\
\hline
Opus & FastAlign & 78.8$_{\pm 0.7}$ & 77.2$_{\pm 0.6}$ & 78.0$_{\pm 0.5}$ & 79.8$_{\pm 0.5}$\\
Opus & AWESOME w/o co & 76.9$_{\pm 1.1}$ & 76.1$_{\pm 1.8}$ & 76.5$_{\pm 1.2}$ & 78.7$_{\pm 1.1}$\\
Opus & AWESOME w/ co & 76.5$_{\pm 1.2}$ & 75.3$_{\pm 1.5}$ & 75.9$_{\pm 1.3}$ & 78.2$_{\pm 1.1}$\\
Opus & AWESOME ft w/o co & 74.6$_{\pm 1.0}$ & 73.9$_{\pm 0.4}$ & 74.2$_{\pm 0.6}$ & 76.8$_{\pm 0.5}$\\
Opus & AWESOME ft w/ co & 74.9$_{\pm 0.3}$ & 75.7$_{\pm 0.8}$ & 75.3$_{\pm 0.3}$ & 78.1$_{\pm 0.5}$\\
Opus & AWESOME pt+ft w/o co & 75.8$_{\pm 1.4}$ & 75.0$_{\pm 1.0}$ & 75.4$_{\pm 1.0}$ & 78.0$_{\pm 0.7}$\\
Opus & AWESOME pt+ft w/ co & 75.7$_{\pm 2.2}$ & 76.2$_{\pm 1.3}$ & 75.9$_{\pm 1.7}$ & 78.3$_{\pm 1.8}$\\
Opus ft & FastAlign & 76.2$_{\pm 2.0}$ & 76.9$_{\pm 2.5}$ & 76.6$_{\pm 2.1}$ & 78.3$_{\pm 2.0}$\\
Opus ft & AWESOME w/o co & 76.1$_{\pm 1.0}$ & 76.9$_{\pm 0.8}$ & 76.5$_{\pm 0.7}$ & 78.9$_{\pm 0.5}$\\
Opus ft & AWESOME w/ co & 74.6$_{\pm 1.5}$ & 76.0$_{\pm 1.3}$ & 75.3$_{\pm 0.9}$ & 77.9$_{\pm 0.6}$\\
Opus ft & AWESOME ft w/o co & 74.1$_{\pm 1.4}$ & 75.5$_{\pm 0.4}$ & 74.8$_{\pm 0.8}$ & 77.4$_{\pm 0.8}$\\
Opus ft & AWESOME ft w/ co & 75.5$_{\pm 1.9}$ & 75.6$_{\pm 1.2}$ & 75.5$_{\pm 1.4}$ & 78.1$_{\pm 1.2}$\\
Opus ft & AWESOME pt+ft w/o co & 75.1$_{\pm 0.5}$ & 76.0$_{\pm 1.2}$ & 75.5$_{\pm 0.6}$ & 78.1$_{\pm 0.5}$\\
Opus ft & AWESOME pt+ft w/ co & 75.0$_{\pm 1.8}$ & 75.8$_{\pm 0.8}$ & 75.4$_{\pm 1.1}$ & 77.7$_{\pm 1.0}$\\
\hline
\multicolumn{2}{c|}{Cross-lingual Transfer} & 78.2$_{\pm 2.5}$ & 77.6$_{\pm 1.2}$ & 77.9$_{\pm 1.7}$ & 80.0$_{\pm 1.4}$\\
\multicolumn{2}{c|}{Cross-lingual Transfer with realignment}  & \textbf{79.7$_{\pm 1.6}$} & \textbf{77.9$_{\pm 1.8}$} & \textbf{78.8$_{\pm 1.6}$} & \textbf{80.8$_{\pm 1.4}$}\\
\hline
    \end{tabular}
    }
    \caption{Cross-lingual Transfer and \ttrain results in French for multilingual base models.}
    \label{tab:full_french}
\end{table*}

\begin{table*}
\centering
\adjustbox{totalheight=\textheight-2\baselineskip}{
    \begin{tabular}{cc|cccc}
\hline
translation & aligner & precision & recall & micro-f1 & macro-f1\\
\hline
\multicolumn{6}{c}{distilmBERT}\\
\hline
FAIR & FastAlign & 69.0$_{\pm 1.2}$ & 63.9$_{\pm 0.5}$ & 66.3$_{\pm 0.8}$ & 44.2$_{\pm 2.4}$\\
FAIR & AWESOME w/o co & 68.1$_{\pm 1.6}$ & 66.1$_{\pm 1.3}$ & 67.1$_{\pm 1.3}$ & 50.5$_{\pm 3.0}$\\
FAIR & AWESOME w/ co & 68.0$_{\pm 2.0}$ & \textbf{67.4$_{\pm 1.4}$} & 67.7$_{\pm 1.0}$ & 50.6$_{\pm 3.7}$\\
FAIR & AWESOME ft w/o co & 68.5$_{\pm 3.6}$ & 65.0$_{\pm 2.2}$ & 66.6$_{\pm 1.2}$ & 48.3$_{\pm 4.0}$\\
FAIR & AWESOME ft w/ co & 67.6$_{\pm 2.4}$ & 66.1$_{\pm 3.5}$ & 66.7$_{\pm 1.4}$ & 46.4$_{\pm 5.1}$\\
FAIR & AWESOME pt+ft w/o co & 69.6$_{\pm 2.7}$ & 65.9$_{\pm 2.8}$ & 67.6$_{\pm 1.1}$ & 46.8$_{\pm 6.4}$\\
FAIR & AWESOME pt+ft w/ co & 68.8$_{\pm 2.6}$ & 66.6$_{\pm 3.7}$ & 67.5$_{\pm 1.4}$ & 49.2$_{\pm 6.0}$\\
FAIR ft & FastAlign & 70.5$_{\pm 1.9}$ & 65.2$_{\pm 1.4}$ & 67.7$_{\pm 1.3}$ & \textbf{52.2$_{\pm 1.5}$}\\
FAIR ft & AWESOME w/o co & 69.3$_{\pm 2.0}$ & 66.9$_{\pm 1.6}$ & 68.0$_{\pm 1.1}$ & 48.0$_{\pm 3.6}$\\
FAIR ft & AWESOME w/ co & 68.6$_{\pm 2.8}$ & 66.4$_{\pm 2.0}$ & 67.4$_{\pm 1.5}$ & 47.4$_{\pm 4.0}$\\
FAIR ft & AWESOME ft w/o co & 70.9$_{\pm 2.7}$ & 67.2$_{\pm 2.1}$ & 69.0$_{\pm 1.9}$ & 49.8$_{\pm 3.7}$\\
FAIR ft & AWESOME ft w/ co & 71.4$_{\pm 2.0}$ & 67.2$_{\pm 1.8}$ & \textbf{69.2$_{\pm 1.2}$} & 49.0$_{\pm 4.3}$\\
FAIR ft & AWESOME pt+ft w/o co & 69.3$_{\pm 1.8}$ & \textbf{67.4$_{\pm 1.9}$} & 68.3$_{\pm 1.3}$ & 49.2$_{\pm 3.2}$\\
FAIR ft & AWESOME pt+ft w/ co & 70.4$_{\pm 1.9}$ & 67.1$_{\pm 1.6}$ & 68.7$_{\pm 1.2}$ & 49.5$_{\pm 3.1}$\\
\hline
\multicolumn{2}{c|}{Cross-lingual Transfer} & 62.8$_{\pm 2.9}$ & 66.4$_{\pm 2.0}$ & 64.6$_{\pm 2.4}$ & 46.3$_{\pm 3.1}$\\
\multicolumn{2}{c|}{Cross-lingual Transfer with realignment}  & \textbf{72.0$_{\pm 2.3}$} & 64.2$_{\pm 1.4}$ & 67.9$_{\pm 1.5}$ & 46.5$_{\pm 5.2}$\\
\hline
\multicolumn{6}{c}{XLM-R Base}\\
\hline
FAIR & FastAlign & 74.0$_{\pm 2.6}$ & 71.6$_{\pm 1.4}$ & 72.8$_{\pm 1.6}$ & 55.5$_{\pm 5.7}$\\
FAIR & AWESOME w/o co & 73.3$_{\pm 1.8}$ & 72.9$_{\pm 0.8}$ & 73.1$_{\pm 1.3}$ & 53.2$_{\pm 3.8}$\\
FAIR & AWESOME w/ co & 73.9$_{\pm 2.8}$ & 72.8$_{\pm 2.2}$ & 73.3$_{\pm 2.3}$ & 53.1$_{\pm 4.4}$\\
FAIR & AWESOME ft w/o co & 74.1$_{\pm 1.1}$ & 74.1$_{\pm 1.1}$ & 74.1$_{\pm 1.1}$ & 57.5$_{\pm 4.3}$\\
FAIR & AWESOME ft w/ co & 75.4$_{\pm 2.3}$ & 74.1$_{\pm 1.4}$ & 74.8$_{\pm 1.8}$ & 58.0$_{\pm 4.2}$\\
FAIR & AWESOME pt+ft w/o co & 74.6$_{\pm 1.8}$ & 73.6$_{\pm 1.1}$ & 74.1$_{\pm 1.3}$ & 56.3$_{\pm 2.2}$\\
FAIR & AWESOME pt+ft w/ co & 74.9$_{\pm 1.9}$ & 73.8$_{\pm 1.0}$ & 74.4$_{\pm 1.4}$ & 57.5$_{\pm 3.3}$\\
FAIR ft & FastAlign & 74.7$_{\pm 2.2}$ & 72.1$_{\pm 0.6}$ & 73.3$_{\pm 1.3}$ & 54.2$_{\pm 4.6}$\\
FAIR ft & AWESOME w/o co & 76.8$_{\pm 1.6}$ & 72.8$_{\pm 1.0}$ & 74.7$_{\pm 1.0}$ & 53.0$_{\pm 2.7}$\\
FAIR ft & AWESOME w/ co & 75.9$_{\pm 1.7}$ & \textbf{73.8$_{\pm 0.6}$} & 74.8$_{\pm 1.0}$ & 57.5$_{\pm 4.5}$\\
FAIR ft & AWESOME ft w/o co & 77.0$_{\pm 1.5}$ & 72.1$_{\pm 0.6}$ & 74.5$_{\pm 0.8}$ & 51.0$_{\pm 0.9}$\\
FAIR ft & AWESOME ft w/ co & 76.2$_{\pm 1.2}$ & 72.1$_{\pm 1.3}$ & 74.1$_{\pm 1.1}$ & 50.9$_{\pm 1.0}$\\
FAIR ft & AWESOME pt+ft w/o co & 75.5$_{\pm 1.2}$ & 71.9$_{\pm 1.1}$ & 73.7$_{\pm 0.9}$ & 52.1$_{\pm 3.5}$\\
FAIR ft & AWESOME pt+ft w/ co & 75.5$_{\pm 0.9}$ & 72.6$_{\pm 1.1}$ & 74.0$_{\pm 1.0}$ & 52.6$_{\pm 3.8}$\\
\hline
\multicolumn{2}{c|}{Cross-lingual Transfer} & 71.1$_{\pm 1.1}$ & 73.3$_{\pm 1.0}$ & 72.2$_{\pm 0.7}$ & 55.1$_{\pm 5.7}$\\
\multicolumn{2}{c|}{Cross-lingual Transfer with realignment}  & \textbf{78.2$_{\pm 1.8}$} & 73.5$_{\pm 1.9}$ & \textbf{75.8$_{\pm 1.3}$} & \textbf{58.1$_{\pm 4.9}$}\\
\hline
\multicolumn{6}{c}{XLM-R Large}\\
\hline
FAIR & FastAlign & 77.7$_{\pm 3.6}$ & 75.1$_{\pm 2.3}$ & 76.4$_{\pm 2.7}$ & 65.8$_{\pm 2.8}$\\
FAIR & AWESOME w/o co & 80.1$_{\pm 1.1}$ & 77.5$_{\pm 0.6}$ & 78.7$_{\pm 0.5}$ & 65.0$_{\pm 3.1}$\\
FAIR & AWESOME w/ co & 79.9$_{\pm 1.3}$ & 77.0$_{\pm 2.4}$ & 78.4$_{\pm 1.7}$ & 64.4$_{\pm 2.2}$\\
FAIR & AWESOME ft w/o co & 80.7$_{\pm 1.6}$ & 77.3$_{\pm 0.9}$ & 79.0$_{\pm 0.6}$ & 65.8$_{\pm 0.6}$\\
FAIR & AWESOME ft w/ co & 80.0$_{\pm 1.7}$ & \textbf{78.5$_{\pm 1.1}$} & 79.2$_{\pm 1.1}$ & 66.3$_{\pm 1.1}$\\
FAIR & AWESOME pt+ft w/o co & 79.3$_{\pm 0.8}$ & 77.3$_{\pm 1.1}$ & 78.3$_{\pm 0.7}$ & 64.5$_{\pm 2.0}$\\
FAIR & AWESOME pt+ft w/ co & 79.1$_{\pm 1.8}$ & 75.6$_{\pm 1.2}$ & 77.3$_{\pm 0.8}$ & 63.7$_{\pm 1.6}$\\
FAIR ft & FastAlign & 78.5$_{\pm 2.5}$ & 75.1$_{\pm 1.6}$ & 76.7$_{\pm 1.7}$ & 61.7$_{\pm 1.9}$\\
FAIR ft & AWESOME w/o co & \textbf{83.2$_{\pm 2.8}$} & 76.0$_{\pm 0.9}$ & \textbf{79.4$_{\pm 1.6}$} & 64.3$_{\pm 2.2}$\\
FAIR ft & AWESOME w/ co & 83.0$_{\pm 1.0}$ & 76.1$_{\pm 1.7}$ & \textbf{79.4$_{\pm 1.3}$} & 64.7$_{\pm 1.3}$\\
FAIR ft & AWESOME ft w/o co & 80.0$_{\pm 1.5}$ & 76.0$_{\pm 1.5}$ & 77.9$_{\pm 1.2}$ & 64.9$_{\pm 0.9}$\\
FAIR ft & AWESOME ft w/ co & 81.7$_{\pm 2.0}$ & 75.1$_{\pm 1.7}$ & 78.3$_{\pm 1.8}$ & 64.8$_{\pm 1.6}$\\
FAIR ft & AWESOME pt+ft w/o co & 80.3$_{\pm 1.9}$ & 74.6$_{\pm 1.6}$ & 77.4$_{\pm 1.3}$ & 62.9$_{\pm 3.2}$\\
FAIR ft & AWESOME pt+ft w/ co & 80.2$_{\pm 0.9}$ & 75.3$_{\pm 2.0}$ & 77.6$_{\pm 1.1}$ & 63.0$_{\pm 2.9}$\\
\hline
\multicolumn{2}{c|}{Cross-lingual Transfer} & 81.2$_{\pm 1.2}$ & 76.0$_{\pm 0.9}$ & 78.5$_{\pm 0.4}$ & 64.9$_{\pm 2.5}$\\
\multicolumn{2}{c|}{Cross-lingual Transfer with realignment}  & 80.7$_{\pm 2.1}$ & 76.0$_{\pm 1.4}$ & 78.3$_{\pm 1.6}$ & \textbf{66.8$_{\pm 1.8}$}\\
\hline
    \end{tabular}
    }
    \caption{Cross-lingual Transfer and \ttrain results in German for multilingual base models.}
    \label{tab:full_german}
\end{table*}

\begin{table*}
\centering
    \begin{tabular}{cc|cccc}
\hline
translation & aligner & precision & recall & mirco-f1 & macro-f1 \\
\hline
\multicolumn{6}{c}{CamemBERT Base}\\
\hline
Opus & FastAlign & \textbf{74.9$_{\pm 1.1}$} & \textbf{78.5$_{\pm 0.9}$} & \textbf{76.7$_{\pm 0.9}$} & \textbf{78.7$_{\pm 1.0}$}\\
Opus & AWESOME w/o co & 73.2$_{\pm 1.2}$ & 77.7$_{\pm 1.3}$ & 75.4$_{\pm 1.2}$ & 77.9$_{\pm 1.2}$\\
Opus & AWESOME w/ co & 74.4$_{\pm 0.8}$ & 77.5$_{\pm 0.8}$ & 75.9$_{\pm 0.8}$ & 78.1$_{\pm 1.0}$\\
Opus & AWESOME ft w/o co & 71.9$_{\pm 1.1}$ & 76.5$_{\pm 1.1}$ & 74.1$_{\pm 1.1}$ & 76.7$_{\pm 1.0}$\\
Opus & AWESOME ft w/ co & 72.3$_{\pm 2.0}$ & 77.4$_{\pm 1.3}$ & 74.8$_{\pm 1.7}$ & 77.3$_{\pm 1.3}$\\
Opus & AWESOME pt+ft w/o co & 74.0$_{\pm 1.4}$ & 77.8$_{\pm 1.5}$ & 75.9$_{\pm 1.4}$ & 78.2$_{\pm 1.3}$\\
Opus & AWESOME pt+ft w/ co & 73.3$_{\pm 1.0}$ & 77.9$_{\pm 1.0}$ & 75.5$_{\pm 0.9}$ & 77.8$_{\pm 1.0}$\\
Opus ft & FastAlign & 74.2$_{\pm 2.1}$ & 78.4$_{\pm 1.3}$ & 76.2$_{\pm 1.7}$ & 78.3$_{\pm 1.5}$\\
Opus ft & AWESOME w/o co & 71.0$_{\pm 1.6}$ & 76.2$_{\pm 1.5}$ & 73.5$_{\pm 1.5}$ & 76.1$_{\pm 1.3}$\\
Opus ft & AWESOME w/ co & 72.0$_{\pm 1.8}$ & 77.4$_{\pm 1.7}$ & 74.6$_{\pm 1.7}$ & 77.2$_{\pm 1.6}$\\
Opus ft & AWESOME ft w/o co & 70.7$_{\pm 1.8}$ & 75.0$_{\pm 1.4}$ & 72.8$_{\pm 1.6}$ & 75.4$_{\pm 1.5}$\\
Opus ft & AWESOME ft w/ co & 72.3$_{\pm 1.6}$ & 76.6$_{\pm 1.6}$ & 74.4$_{\pm 1.6}$ & 76.8$_{\pm 1.5}$\\
Opus ft & AWESOME pt+ft w/o co & 72.2$_{\pm 2.4}$ & 77.1$_{\pm 1.5}$ & 74.6$_{\pm 1.9}$ & 76.9$_{\pm 1.7}$\\
Opus ft & AWESOME pt+ft w/ co & 71.6$_{\pm 1.2}$ & 77.0$_{\pm 0.4}$ & 74.2$_{\pm 0.8}$ & 76.7$_{\pm 0.4}$\\
\hline
\multicolumn{6}{c}{DrBERT 7GB}\\
\hline
Opus & FastAlign & 70.9$_{\pm 2.4}$ & 72.4$_{\pm 2.1}$ & 71.7$_{\pm 2.1}$ & 73.2$_{\pm 2.1}$\\
Opus & AWESOME w/o co & 69.5$_{\pm 1.4}$ & 71.3$_{\pm 1.5}$ & 70.4$_{\pm 1.3}$ & 72.3$_{\pm 1.6}$\\
Opus & AWESOME w/ co & 69.5$_{\pm 1.7}$ & 71.7$_{\pm 0.7}$ & 70.6$_{\pm 0.8}$ & 72.6$_{\pm 0.7}$\\
Opus & AWESOME ft w/o co & 69.3$_{\pm 1.1}$ & 71.7$_{\pm 0.7}$ & 70.4$_{\pm 0.8}$ & 72.7$_{\pm 0.7}$\\
Opus & AWESOME ft w/ co & 68.1$_{\pm 0.8}$ & 70.3$_{\pm 1.5}$ & 69.2$_{\pm 0.7}$ & 71.3$_{\pm 0.8}$\\
Opus & AWESOME pt+ft w/o co & 69.4$_{\pm 1.4}$ & 71.7$_{\pm 1.3}$ & 70.5$_{\pm 1.2}$ & 72.6$_{\pm 1.1}$\\
Opus & AWESOME pt+ft w/ co & 70.0$_{\pm 1.4}$ & 71.4$_{\pm 1.0}$ & 70.7$_{\pm 0.6}$ & 72.7$_{\pm 0.5}$\\
Opus ft & FastAlign & \textbf{73.2$_{\pm 1.9}$} &\textbf{ 73.7$_{\pm 1.5}$} & \textbf{73.5$_{\pm 1.4}$} & \textbf{74.9$_{\pm 1.4}$}\\
Opus ft & AWESOME w/o co & 69.6$_{\pm 1.6}$ & 71.7$_{\pm 1.1}$ & 70.7$_{\pm 1.3}$ & 72.7$_{\pm 1.1}$\\
Opus ft & AWESOME w/ co & 70.5$_{\pm 1.6}$ & 71.9$_{\pm 1.2}$ & 71.2$_{\pm 1.2}$ & 73.4$_{\pm 1.1}$\\
Opus ft & AWESOME ft w/o co & 69.1$_{\pm 1.5}$ & 70.6$_{\pm 1.1}$ & 69.8$_{\pm 0.5}$ & 71.8$_{\pm 0.3}$\\
Opus ft & AWESOME ft w/ co & 70.8$_{\pm 1.5}$ & 72.6$_{\pm 1.3}$ & 71.6$_{\pm 1.0}$ & 73.7$_{\pm 0.8}$\\
Opus ft & AWESOME pt+ft w/o co & 70.7$_{\pm 1.0}$ & 71.7$_{\pm 2.1}$ & 71.2$_{\pm 1.4}$ & 73.2$_{\pm 1.2}$\\
Opus ft & AWESOME pt+ft w/ co & 70.1$_{\pm 0.8}$ & 70.6$_{\pm 1.7}$ & 70.4$_{\pm 1.1}$ & 72.4$_{\pm 1.2}$\\
\hline
\multicolumn{6}{c}{DrBERT-PubMedBERT}\\
\hline
Opus & FastAlign & \textbf{76.2$_{\pm 1.4}$} & 79.4$_{\pm 0.8}$ & 77.8$_{\pm 1.1}$ & 79.7$_{\pm 0.8}$\\
Opus & AWESOME w/o co & 75.2$_{\pm 1.2}$ & 77.5$_{\pm 0.7}$ & 76.4$_{\pm 0.9}$ & 78.4$_{\pm 0.7}$\\
Opus & AWESOME w/ co & 76.0$_{\pm 1.0}$ & 79.2$_{\pm 1.3}$ & 77.6$_{\pm 1.1}$ & 79.6$_{\pm 0.9}$\\
Opus & AWESOME ft w/o co & 74.3$_{\pm 1.4}$ & 78.9$_{\pm 1.3}$ & 76.5$_{\pm 1.2}$ & 78.7$_{\pm 1.1}$\\
Opus & AWESOME ft w/ co & 74.0$_{\pm 1.2}$ & 77.8$_{\pm 1.1}$ & 75.9$_{\pm 1.1}$ & 78.2$_{\pm 1.0}$\\
Opus & AWESOME pt+ft w/o co & 73.9$_{\pm 1.1}$ & 77.7$_{\pm 0.6}$ & 75.8$_{\pm 0.9}$ & 78.1$_{\pm 0.7}$\\
Opus & AWESOME pt+ft w/ co & 75.2$_{\pm 0.4}$ & 79.1$_{\pm 0.8}$ & 77.1$_{\pm 0.5}$ & 79.0$_{\pm 0.4}$\\
Opus ft & FastAlign & \textbf{76.2$_{\pm 1.8}$} & \textbf{81.5$_{\pm 1.0}$} & \textbf{78.8$_{\pm 1.4}$} & \textbf{80.4$_{\pm 1.3}$}\\
Opus ft & AWESOME w/o co & 73.7$_{\pm 1.4}$ & 78.7$_{\pm 1.2}$ & 76.1$_{\pm 1.3}$ & 78.4$_{\pm 1.0}$\\
Opus ft & AWESOME w/ co & 75.4$_{\pm 1.2}$ & 81.2$_{\pm 0.8}$ & 78.2$_{\pm 0.8}$ & 80.3$_{\pm 0.7}$\\
Opus ft & AWESOME ft w/o co & 74.9$_{\pm 0.9}$ & 80.7$_{\pm 0.7}$ & 77.7$_{\pm 0.8}$ & 79.7$_{\pm 0.5}$\\
Opus ft & AWESOME ft w/ co & 74.8$_{\pm 1.3}$ & 79.7$_{\pm 0.8}$ & 77.2$_{\pm 1.1}$ & 79.2$_{\pm 0.8}$\\
Opus ft & AWESOME pt+ft w/o co & 75.6$_{\pm 1.1}$ & 79.3$_{\pm 1.5}$ & 77.4$_{\pm 1.2}$ & 79.4$_{\pm 1.1}$\\
Opus ft & AWESOME pt+ft w/ co & 75.5$_{\pm 1.2}$ & 80.3$_{\pm 1.3}$ & 77.8$_{\pm 1.2}$ & 79.8$_{\pm 1.0}$\\
\hline
\end{tabular}
\caption{\ttrain results in French for domain and language-specific base models.}
    \label{tab:full_specific_french}
\end{table*}

\begin{table*}
    \centering
    \begin{tabular}{cc|cccc}
\hline
translation & aligner & precision & recall & micro-f1 & macro-f1\\
\hline
\multicolumn{6}{c}{GottBERT}\\
\hline
FAIR & FastAlign & 75.9$_{\pm 3.2}$ & 70.3$_{\pm 2.2}$ & 73.0$_{\pm 2.6}$ & 54.8$_{\pm 4.8}$\\
FAIR & AWESOME w/o co & 79.5$_{\pm 1.6}$ & 73.4$_{\pm 2.0}$ & 76.3$_{\pm 1.7}$ & 60.5$_{\pm 4.6}$\\
FAIR & AWESOME w/ co & 77.9$_{\pm 1.6}$ & 72.9$_{\pm 1.4}$ & 75.3$_{\pm 1.4}$ & 57.2$_{\pm 5.2}$\\
FAIR & AWESOME ft w/o co & 78.4$_{\pm 2.8}$ & 73.1$_{\pm 2.4}$ & 75.7$_{\pm 2.4}$ & 57.9$_{\pm 5.4}$\\
FAIR & AWESOME ft w/ co & 77.9$_{\pm 1.8}$ & 72.6$_{\pm 2.3}$ & 75.1$_{\pm 1.8}$ & 53.4$_{\pm 2.7}$\\
FAIR & AWESOME pt+ft w/o co & 79.0$_{\pm 1.8}$ & \textbf{74.1$_{\pm 1.6}$} & 76.5$_{\pm 1.6}$ & \textbf{60.8$_{\pm 3.5}$}\\
FAIR & AWESOME pt+ft w/ co & 77.9$_{\pm 2.8}$ & 73.6$_{\pm 2.5}$ & 75.7$_{\pm 2.6}$ & 57.6$_{\pm 4.4}$\\
FAIR ft & FastAlign & 76.6$_{\pm 3.2}$ & 70.1$_{\pm 1.9}$ & 73.2$_{\pm 2.4}$ & 53.7$_{\pm 4.5}$\\
FAIR ft & AWESOME w/o co & \textbf{80.2$_{\pm 1.1}$} & 73.3$_{\pm 1.0}$ & \textbf{76.6$_{\pm 0.8}$} & 58.7$_{\pm 6.1}$\\
FAIR ft & AWESOME w/ co & 79.2$_{\pm 0.8}$ & 73.1$_{\pm 1.1}$ & 76.0$_{\pm 0.7}$ & 58.8$_{\pm 2.4}$\\
FAIR ft & AWESOME ft w/o co & 78.5$_{\pm 0.7}$ & 72.4$_{\pm 1.0}$ & 75.3$_{\pm 0.8}$ & 56.1$_{\pm 6.6}$\\
FAIR ft & AWESOME ft w/ co & 78.8$_{\pm 2.3}$ & 72.3$_{\pm 1.9}$ & 75.4$_{\pm 1.8}$ & 55.2$_{\pm 6.4}$\\
FAIR ft & AWESOME pt+ft w/o co & 78.6$_{\pm 1.6}$ & 72.6$_{\pm 1.6}$ & 75.5$_{\pm 1.4}$ & 55.5$_{\pm 6.4}$\\
FAIR ft & AWESOME pt+ft w/ co & 79.4$_{\pm 1.8}$ & 72.3$_{\pm 0.8}$ & 75.6$_{\pm 1.1}$ & 56.5$_{\pm 3.2}$\\
\hline
\multicolumn{6}{c}{medBERT.de}\\
\hline
FAIR & FastAlign & 75.1$_{\pm 2.8}$ & 69.2$_{\pm 1.6}$ & 72.0$_{\pm 2.0}$ & 56.7$_{\pm 5.3}$\\
FAIR & AWESOME w/o co & 75.4$_{\pm 1.6}$ & 71.9$_{\pm 1.8}$ & 73.6$_{\pm 1.2}$ & 58.1$_{\pm 5.0}$\\
FAIR & AWESOME w/ co & \textbf{77.0$_{\pm 3.5}$} & 72.9$_{\pm 1.6}$ & 74.9$_{\pm 2.4}$ & 59.5$_{\pm 5.9}$\\
FAIR & AWESOME ft w/o co & 76.4$_{\pm 4.5}$ & 70.9$_{\pm 1.6}$ & 73.5$_{\pm 2.7}$ & 56.2$_{\pm 5.4}$\\
FAIR & AWESOME ft w/ co & 75.8$_{\pm 4.0}$ & 72.4$_{\pm 2.5}$ & 74.1$_{\pm 3.1}$ & 57.2$_{\pm 6.4}$\\
FAIR & AWESOME pt+ft w/o co & 74.9$_{\pm 3.9}$ & 71.3$_{\pm 1.6}$ & 73.0$_{\pm 2.4}$ & 56.7$_{\pm 5.5}$\\
FAIR & AWESOME pt+ft w/ co & 76.1$_{\pm 4.1}$ & 71.6$_{\pm 1.9}$ & 73.7$_{\pm 2.6}$ & 57.2$_{\pm 5.7}$\\
FAIR ft & FastAlign & 72.6$_{\pm 2.0}$ & 68.9$_{\pm 0.8}$ & 70.7$_{\pm 1.2}$ & 60.7$_{\pm 1.2}$\\
FAIR ft & AWESOME w/o co & 75.2$_{\pm 2.0}$ & 72.6$_{\pm 0.9}$ & 73.9$_{\pm 1.0}$ & 61.1$_{\pm 1.9}$\\
FAIR ft & AWESOME w/ co & 76.4$_{\pm 2.4}$ & \textbf{73.6$_{\pm 0.9}$} & \textbf{75.0$_{\pm 1.6}$} & 62.2$_{\pm 3.3}$\\
FAIR ft & AWESOME ft w/o co & 75.1$_{\pm 3.6}$ & 71.8$_{\pm 2.8}$ & 73.4$_{\pm 3.1}$ & 60.5$_{\pm 3.5}$\\
FAIR ft & AWESOME ft w/ co & 75.3$_{\pm 3.2}$ & 71.9$_{\pm 2.2}$ & 73.6$_{\pm 2.4}$ & 58.6$_{\pm 6.1}$\\
FAIR ft & AWESOME pt+ft w/o co & 74.1$_{\pm 1.1}$ & 71.4$_{\pm 0.8}$ & 72.7$_{\pm 0.5}$ & 57.8$_{\pm 4.9}$\\
FAIR ft & AWESOME pt+ft w/ co & 75.5$_{\pm 1.3}$ & 72.3$_{\pm 1.0}$ & 73.8$_{\pm 1.0}$ & \textbf{62.6$_{\pm 1.2}$}\\
\hline
    \end{tabular}
\caption{\ttrain results in German for domain and language-specific base models.}
    \label{tab:full_specific_german}
\end{table*}

\begin{table*}
\centering
    \begin{tabular}{cc|cccc}
\hline
translation & aligner & precision & recall & micro-f1 & macro-f1\\
\hline
\multicolumn{6}{c}{distilmBERT}\\
\hline
Opus & FastAlign & 65.9$_{\pm 1.8}$ & 67.0$_{\pm 1.5}$ & 66.4$_{\pm 1.5}$ & 68.5$_{\pm 1.3}$\\
Opus & AWESOME w/o co & \textbf{70.5$_{\pm 1.7}$} & 68.8$_{\pm 1.4}$ & 69.6$_{\pm 1.4}$ & 71.8$_{\pm 1.3}$\\
Opus & AWESOME w/ co & \textbf{70.5$_{\pm 1.7}$} & \textbf{69.0$_{\pm 1.4}$} & \textbf{69.7$_{\pm 1.5}$} & \textbf{71.9$_{\pm 1.3}$}\\
Opus & AWESOME ft w/o co & 70.0$_{\pm 1.8}$ & 68.8$_{\pm 1.6}$ & 69.4$_{\pm 1.6}$ & 71.6$_{\pm 1.4}$\\
Opus & AWESOME ft w/ co & 69.7$_{\pm 1.8}$ & 68.8$_{\pm 1.6}$ & 69.2$_{\pm 1.6}$ & 71.5$_{\pm 1.4}$\\
Opus & AWESOME pt+ft w/o co & 70.1$_{\pm 1.7}$ & 68.9$_{\pm 1.5}$ & 69.5$_{\pm 1.5}$ & 71.7$_{\pm 1.4}$\\
Opus & AWESOME pt+ft w/ co & 69.0$_{\pm 1.8}$ & 68.8$_{\pm 1.6}$ & 68.9$_{\pm 1.6}$ & 71.3$_{\pm 1.4}$\\
Opus ft & FastAlign & 63.2$_{\pm 1.7}$ & 66.6$_{\pm 1.6}$ & 64.8$_{\pm 1.5}$ & 66.7$_{\pm 1.2}$\\
Opus ft & AWESOME w/o co & 69.9$_{\pm 1.6}$ & 68.4$_{\pm 1.3}$ & 69.2$_{\pm 1.4}$ & 71.4$_{\pm 1.2}$\\
Opus ft & AWESOME w/ co & 69.2$_{\pm 1.7}$ & 68.7$_{\pm 1.4}$ & 68.9$_{\pm 1.5}$ & 71.3$_{\pm 1.3}$\\
Opus ft & AWESOME ft w/o co & 69.4$_{\pm 1.8}$ & 68.6$_{\pm 1.6}$ & 69.0$_{\pm 1.6}$ & 71.2$_{\pm 1.4}$\\
Opus ft & AWESOME ft w/ co & 69.1$_{\pm 1.7}$ & 68.6$_{\pm 1.5}$ & 68.9$_{\pm 1.5}$ & 71.1$_{\pm 1.3}$\\
Opus ft & AWESOME pt+ft w/o co & 69.1$_{\pm 1.7}$ & 68.6$_{\pm 1.5}$ & 68.8$_{\pm 1.5}$ & 71.1$_{\pm 1.3}$\\
Opus ft & AWESOME pt+ft w/ co & 67.3$_{\pm 1.7}$ & 68.6$_{\pm 1.5}$ & 67.9$_{\pm 1.5}$ & 70.5$_{\pm 1.3}$\\
\hline
\multicolumn{6}{c}{XLM-R Base}\\
\hline
Opus & FastAlign & 68.7$_{\pm 1.7}$ & 71.8$_{\pm 1.2}$ & 70.2$_{\pm 1.4}$ & 72.4$_{\pm 1.2}$\\
Opus & AWESOME w/o co & 74.9$_{\pm 1.5}$ & 73.8$_{\pm 1.2}$ & 74.3$_{\pm 1.3}$ & 76.5$_{\pm 1.1}$\\
Opus & AWESOME w/ co & 74.8$_{\pm 1.6}$ & \textbf{74.1$_{\pm 1.3}$} & 7\textbf{4.4$_{\pm 1.3}$} & \textbf{76.6$_{\pm 1.2}$}\\
Opus & AWESOME ft w/o co & 74.6$_{\pm 1.6}$ & 73.8$_{\pm 1.2}$ & 74.2$_{\pm 1.2}$ & 76.3$_{\pm 1.1}$\\
Opus & AWESOME ft w/ co & 74.1$_{\pm 1.5}$ & 73.8$_{\pm 1.2}$ & 74.0$_{\pm 1.3}$ & 76.1$_{\pm 1.1}$\\
Opus & AWESOME pt+ft w/o co & 74.6$_{\pm 1.6}$ & 73.8$_{\pm 1.2}$ & 74.2$_{\pm 1.2}$ & 76.3$_{\pm 1.1}$\\
Opus & AWESOME pt+ft w/ co & 73.4$_{\pm 1.5}$ & 73.8$_{\pm 1.2}$ & 73.6$_{\pm 1.2}$ & 75.9$_{\pm 1.1}$\\
Opus ft & FastAlign & 67.7$_{\pm 1.6}$ & 71.7$_{\pm 1.7}$ & 69.6$_{\pm 1.4}$ & 71.4$_{\pm 1.3}$\\
Opus ft & AWESOME w/o co & \textbf{75.4$_{\pm 1.7}$} & 73.1$_{\pm 1.7}$ & 74.2$_{\pm 1.6}$ & 76.3$_{\pm 1.5}$\\
Opus ft & AWESOME w/ co & 74.2$_{\pm 1.7}$ & 73.3$_{\pm 1.7}$ & 73.8$_{\pm 1.6}$ & 76.0$_{\pm 1.6}$\\
Opus ft & AWESOME ft w/o co & 74.6$_{\pm 1.8}$ & 73.1$_{\pm 1.7}$ & 73.8$_{\pm 1.6}$ & 75.8$_{\pm 1.5}$\\
Opus ft & AWESOME ft w/ co & 74.1$_{\pm 1.8}$ & 73.1$_{\pm 1.7}$ & 73.6$_{\pm 1.6}$ & 75.6$_{\pm 1.5}$\\
Opus ft & AWESOME pt+ft w/o co & 74.3$_{\pm 1.8}$ & 73.2$_{\pm 1.8}$ & 73.7$_{\pm 1.7}$ & 75.7$_{\pm 1.6}$\\
Opus ft & AWESOME pt+ft w/ co & 72.3$_{\pm 1.7}$ & 73.1$_{\pm 1.7}$ & 72.7$_{\pm 1.6}$ & 75.0$_{\pm 1.5}$\\
\hline
\multicolumn{6}{c}{XLM-R Large}\\
\hline
Opus & FastAlign & 69.6$_{\pm 1.3}$ & 71.0$_{\pm 0.9}$ & 70.3$_{\pm 1.1}$ & 72.7$_{\pm 0.9}$\\
Opus & AWESOME w/o co & 75.9$_{\pm 1.0}$ & 73.4$_{\pm 0.8}$ & 74.7$_{\pm 0.9}$ & 77.0$_{\pm 0.8}$\\
Opus & AWESOME w/ co & 76.0$_{\pm 1.0}$ & \textbf{73.7$_{\pm 0.7}$} & 74.8$_{\pm 0.8}$ & 77.2$_{\pm 0.8}$\\
Opus & AWESOME ft w/o co & 75.5$_{\pm 1.0}$ & 73.5$_{\pm 0.8}$ & 74.5$_{\pm 0.9}$ & 76.8$_{\pm 0.8}$\\
Opus & AWESOME ft w/ co & 75.2$_{\pm 1.1}$ & 73.5$_{\pm 0.8}$ & 74.3$_{\pm 0.9}$ & 76.7$_{\pm 0.8}$\\
Opus & AWESOME pt+ft w/o co & 75.6$_{\pm 1.1}$ & 73.5$_{\pm 0.8}$ & 74.5$_{\pm 0.9}$ & 76.8$_{\pm 0.8}$\\
Opus & AWESOME pt+ft w/ co & 74.5$_{\pm 1.1}$ & 73.5$_{\pm 0.8}$ & 74.0$_{\pm 0.9}$ & 76.5$_{\pm 0.8}$\\
Opus ft & FastAlign & 70.8$_{\pm 1.4}$ & 72.2$_{\pm 0.5}$ & 71.5$_{\pm 1.0}$ & 73.2$_{\pm 0.9}$\\
Opus ft & AWESOME w/o co & \textbf{77.3$_{\pm 1.3}$} & 73.3$_{\pm 0.6}$ & \textbf{75.3$_{\pm 0.9}$} & \textbf{77.6$_{\pm 0.7}$}\\
Opus ft & AWESOME w/ co & 76.3$_{\pm 1.1}$ & 73.6$_{\pm 0.6}$ & 75.0$_{\pm 0.9}$ & 77.4$_{\pm 0.6}$\\
Opus ft & AWESOME ft w/o co & 76.4$_{\pm 1.3}$ & 73.4$_{\pm 0.6}$ & 74.9$_{\pm 0.9}$ & 77.1$_{\pm 0.7}$\\
Opus ft & AWESOME ft w/ co & 76.1$_{\pm 1.2}$ & 73.4$_{\pm 0.6}$ & 74.7$_{\pm 0.9}$ & 77.0$_{\pm 0.7}$\\
Opus ft & AWESOME pt+ft w/o co & 76.1$_{\pm 1.2}$ & 73.5$_{\pm 0.6}$ & 74.8$_{\pm 0.8}$ & 77.0$_{\pm 0.7}$\\
Opus ft & AWESOME pt+ft w/ co & 74.1$_{\pm 1.1}$ & 73.4$_{\pm 0.6}$ & 73.8$_{\pm 0.8}$ & 76.3$_{\pm 0.6}$\\
\hline
\end{tabular}
\caption{Full results for the \ttest approach in French with multilingual language models.}
\label{tab:full_test_fr}
\end{table*}

\begin{table*}
\centering
    \begin{tabular}{cc|cccc}
\hline
translation & aligner & precision & recall & micro-f1 & macro-f1\\
\hline
\multicolumn{6}{c}{distilmBERT}\\
\hline
FAIR & FastAlign & 37.6$_{\pm 2.2}$ & 36.5$_{\pm 1.1}$ & 37.0$_{\pm 1.5}$ & 24.4$_{\pm 1.2}$\\
FAIR & AWESOME w/o co & 66.8$_{\pm 2.7}$ & 63.7$_{\pm 1.4}$ & 65.2$_{\pm 1.9}$ & 49.1$_{\pm 1.6}$\\
FAIR & AWESOME w/ co & 66.9$_{\pm 2.7}$ & 63.9$_{\pm 1.6}$ & 65.3$_{\pm 1.9}$ & 49.3$_{\pm 1.7}$\\
FAIR & AWESOME ft w/o co & 68.6$_{\pm 2.9}$ & 63.7$_{\pm 1.4}$ & 66.0$_{\pm 1.9}$ & 49.6$_{\pm 1.6}$\\
FAIR & AWESOME ft w/ co & 68.8$_{\pm 2.8}$ & 64.5$_{\pm 1.4}$ & 66.6$_{\pm 1.9}$ & 50.2$_{\pm 1.6}$\\
FAIR & AWESOME pt+ft w/o co & 67.2$_{\pm 2.8}$ & 62.9$_{\pm 1.4}$ & 64.9$_{\pm 1.9}$ & 48.6$_{\pm 1.6}$\\
FAIR & AWESOME pt+ft w/ co & 67.3$_{\pm 2.9}$ & 63.9$_{\pm 1.6}$ & 65.5$_{\pm 2.0}$ & 49.4$_{\pm 1.7}$\\
FAIR ft & FastAlign & 29.1$_{\pm 0.6}$ & 31.1$_{\pm 0.8}$ & 30.1$_{\pm 0.6}$ & 18.3$_{\pm 0.6}$\\
FAIR ft & AWESOME w/o co & 69.8$_{\pm 2.6}$ & 65.4$_{\pm 1.4}$ & 67.5$_{\pm 1.8}$ & 50.9$_{\pm 3.8}$\\
FAIR ft & AWESOME w/ co & 68.9$_{\pm 2.5}$ & 66.4$_{\pm 1.4}$ & 67.6$_{\pm 1.7}$ & 51.4$_{\pm 3.8}$\\
FAIR ft & AWESOME ft w/o co & 69.5$_{\pm 2.3}$ & 65.4$_{\pm 1.4}$ & 67.4$_{\pm 1.7}$ & 50.8$_{\pm 3.8}$\\
FAIR ft & AWESOME ft w/ co & 69.7$_{\pm 2.3}$ & 65.5$_{\pm 1.5}$ & 67.5$_{\pm 1.6}$ & 51.0$_{\pm 3.8}$\\
FAIR ft & AWESOME pt+ft w/o co & \textbf{69.9$_{\pm 2.4}$} & \textbf{66.7$_{\pm 1.6}$} & \textbf{68.3$_{\pm 1.8}$} & \textbf{51.9$_{\pm 3.7}$}\\
FAIR ft & AWESOME pt+ft w/ co & 69.2$_{\pm 2.1}$ & 66.2$_{\pm 1.4}$ & 67.6$_{\pm 1.6}$ & 51.3$_{\pm 3.8}$\\
\hline
\multicolumn{6}{c}{XLM-R Base}\\
\hline
FAIR & FastAlign & 37.0$_{\pm 1.5}$ & 41.2$_{\pm 0.8}$ & 39.0$_{\pm 0.8}$ & 28.4$_{\pm 0.5}$\\
FAIR & AWESOME w/o co & 71.1$_{\pm 0.7}$ & 72.3$_{\pm 0.8}$ & 71.7$_{\pm 0.3}$ & 56.4$_{\pm 0.6}$\\
FAIR & AWESOME w/ co & 71.1$_{\pm 0.7}$ & 72.3$_{\pm 0.8}$ & 71.7$_{\pm 0.3}$ & 56.4$_{\pm 0.6}$\\
FAIR & AWESOME ft w/o co & 72.0$_{\pm 0.7}$ & 71.4$_{\pm 0.8}$ & 71.7$_{\pm 0.3}$ & 56.1$_{\pm 0.7}$\\
FAIR & AWESOME ft w/ co & \textbf{72.3$_{\pm 0.7}$} & 72.3$_{\pm 0.8}$ & 72.3$_{\pm 0.3}$ & \textbf{56.7$_{\pm 0.7}$}\\
FAIR & AWESOME pt+ft w/o co & 70.6$_{\pm 0.7}$ & 70.6$_{\pm 0.8}$ & 70.6$_{\pm 0.3}$ & 55.2$_{\pm 0.6}$\\
FAIR & AWESOME pt+ft w/ co & 70.8$_{\pm 0.7}$ & 71.4$_{\pm 0.8}$ & 71.1$_{\pm 0.3}$ & 55.8$_{\pm 0.6}$\\
FAIR ft & FastAlign & 29.7$_{\pm 0.7}$ & 35.1$_{\pm 1.1}$ & 32.2$_{\pm 0.7}$ & 22.0$_{\pm 0.8}$\\
FAIR ft & AWESOME w/o co & 71.0$_{\pm 1.2}$ & 71.9$_{\pm 2.0}$ & 71.4$_{\pm 1.2}$ & 54.3$_{\pm 5.2}$\\
FAIR ft & AWESOME w/ co & 70.4$_{\pm 0.8}$ & 72.9$_{\pm 1.6}$ & 71.6$_{\pm 0.6}$ & 54.8$_{\pm 5.0}$\\
FAIR ft & AWESOME ft w/o co & 71.7$_{\pm 1.2}$ & 72.4$_{\pm 1.6}$ & 72.1$_{\pm 0.8}$ & 55.1$_{\pm 5.3}$\\
FAIR ft & AWESOME ft w/ co & 72.1$_{\pm 1.2}$ & 72.4$_{\pm 1.6}$ & 72.3$_{\pm 0.7}$ & 55.2$_{\pm 5.2}$\\
FAIR ft & AWESOME pt+ft w/o co & 72.0$_{\pm 1.0}$ & \textbf{73.4$_{\pm 1.7}$} & \textbf{72.7$_{\pm 0.8}$} & 55.5$_{\pm 5.0}$\\
FAIR ft & AWESOME pt+ft w/ co & 71.7$_{\pm 1.1}$ & 72.8$_{\pm 2.0}$ & 72.2$_{\pm 1.0}$ & 55.3$_{\pm 5.5}$\\
\hline
\multicolumn{6}{c}{XLM-R Large}\\
\hline
FAIR & FastAlign & 39.9$_{\pm 1.7}$ & 41.4$_{\pm 0.7}$ & 40.6$_{\pm 1.0}$ & 28.7$_{\pm 1.2}$\\
FAIR & AWESOME w/o co & 76.4$_{\pm 1.6}$ & 72.5$_{\pm 0.9}$ & 74.4$_{\pm 1.0}$ & 49.2$_{\pm 0.4}$\\
FAIR & AWESOME w/ co & 76.4$_{\pm 1.6}$ & 72.5$_{\pm 0.9}$ & 74.4$_{\pm 1.0}$ & 49.2$_{\pm 0.4}$\\
FAIR & AWESOME ft w/o co & 77.5$_{\pm 1.7}$ & 71.6$_{\pm 0.9}$ & 74.5$_{\pm 1.0}$ & 49.0$_{\pm 0.5}$\\
FAIR & AWESOME ft w/ co & 77.7$_{\pm 1.7}$ & 72.5$_{\pm 0.9}$ & 75.0$_{\pm 1.0}$ & 49.6$_{\pm 0.5}$\\
FAIR & AWESOME pt+ft w/o co & 75.9$_{\pm 1.7}$ & 70.8$_{\pm 0.9}$ & 73.3$_{\pm 1.0}$ & 47.9$_{\pm 0.4}$\\
FAIR & AWESOME pt+ft w/ co & 76.1$_{\pm 1.6}$ & 71.6$_{\pm 0.9}$ & 73.8$_{\pm 1.0}$ & 48.6$_{\pm 0.4}$\\
FAIR ft & FastAlign & 30.6$_{\pm 1.5}$ & 34.0$_{\pm 0.9}$ & 32.2$_{\pm 1.2}$ & 20.9$_{\pm 0.9}$\\
FAIR ft & AWESOME w/o co & 76.6$_{\pm 3.5}$ & 72.7$_{\pm 2.1}$ & 74.6$_{\pm 2.7}$ & 49.4$_{\pm 1.8}$\\
FAIR ft & AWESOME w/ co & 76.5$_{\pm 3.7}$ & \textbf{74.4$_{\pm 2.1}$} & 75.4$_{\pm 2.8}$ & 50.7$_{\pm 1.9}$\\
FAIR ft & AWESOME ft w/o co & 77.8$_{\pm 3.6}$ & 73.3$_{\pm 2.2}$ & 75.5$_{\pm 2.7}$ & 50.1$_{\pm 1.9}$\\
FAIR ft & AWESOME ft w/ co & \textbf{78.3$_{\pm 3.7}$} & 73.3$_{\pm 2.2}$ & 75.7$_{\pm 2.8}$ & 50.2$_{\pm 1.9}$\\
FAIR ft & AWESOME pt+ft w/o co & 77.9$_{\pm 3.8}$ & \textbf{74.4$_{\pm 2.1}$} & \textbf{76.1$_{\pm 2.8}$} & \textbf{51.0$_{\pm 1.9}$}\\
FAIR ft & AWESOME pt+ft w/ co & 77.5$_{\pm 3.8}$ & 73.5$_{\pm 2.1}$ & 75.5$_{\pm 2.8}$ & 50.2$_{\pm 1.9}$\\
\hline
\end{tabular}
\caption{Full results for the \ttest approach in German with multilingual language models.}
\label{tab:full_test_de}
\end{table*}

\begin{table*}
    \centering
    \begin{tabular}{cc|cccc}
    \hline
translation & aligner & precision & recall & micro-f1 & macro-f1\\
\hline
Opus & FastAlign & 68.5$_{\pm 1.2}$ & 69.1$_{\pm 1.2}$ & 68.8$_{\pm 1.0}$ & 71.2$_{\pm 1.0}$\\
Opus & AWESOME w/o co & 75.2$_{\pm 1.4}$ & \textbf{71.8$_{\pm 1.3}$} & \textbf{73.5$_{\pm 1.2}$} & \textbf{75.8$_{\pm 1.2}$}\\
Opus & AWESOME w/ co & 74.8$_{\pm 1.5}$ & \textbf{71.8$_{\pm 1.4}$} & 73.3$_{\pm 1.3}$ & 75.6$_{\pm 1.2}$\\
Opus & AWESOME ft w/o co & 74.7$_{\pm 1.4}$ & \textbf{71.8$_{\pm 1.3}$} & 73.2$_{\pm 1.2}$ & 75.5$_{\pm 1.2}$\\
Opus & AWESOME ft w/ co & 74.1$_{\pm 1.4}$ & 71.5$_{\pm 1.3}$ & 72.8$_{\pm 1.2}$ & 75.1$_{\pm 1.2}$\\
Opus & AWESOME pt+ft w/o co & 74.7$_{\pm 1.4}$ & \textbf{71.8$_{\pm 1.3}$} & 73.2$_{\pm 1.2}$ & 75.5$_{\pm 1.2}$\\
Opus & AWESOME pt+ft w/ co & 73.3$_{\pm 1.4}$ & 71.5$_{\pm 1.3}$ & 72.4$_{\pm 1.2}$ & 74.9$_{\pm 1.2}$\\
Opus ft & FastAlign & 67.9$_{\pm 1.3}$ & 69.3$_{\pm 1.5}$ & 68.6$_{\pm 1.2}$ & 70.6$_{\pm 1.2}$\\
Opus ft & AWESOME w/o co & \textbf{75.4$_{\pm 1.3}$} & 71.4$_{\pm 1.6}$ & 73.3$_{\pm 1.3}$ & \textbf{75.8$_{\pm 1.3}$}\\
Opus ft & AWESOME w/ co & 74.1$_{\pm 1.4}$ & 71.4$_{\pm 1.7}$ & 72.7$_{\pm 1.4}$ & 75.4$_{\pm 1.4}$\\
Opus ft & AWESOME ft w/o co & 74.7$_{\pm 1.4}$ & 71.4$_{\pm 1.5}$ & 73.0$_{\pm 1.3}$ & 75.4$_{\pm 1.3}$\\
Opus ft & AWESOME ft w/ co & 74.3$_{\pm 1.4}$ & 71.5$_{\pm 1.6}$ & 72.9$_{\pm 1.4}$ & 75.3$_{\pm 1.4}$\\
Opus ft & AWESOME pt+ft w/o co & 74.3$_{\pm 1.3}$ & 71.5$_{\pm 1.6}$ & 72.9$_{\pm 1.3}$ & 75.4$_{\pm 1.3}$\\
Opus ft & AWESOME pt+ft w/ co & 72.3$_{\pm 1.3}$ & 71.5$_{\pm 1.6}$ & 71.9$_{\pm 1.3}$ & 74.7$_{\pm 1.3}$\\
\hline
    \end{tabular}
    \caption{Results of \ttest in French with PubMedBERT.}
    \label{tab:pubmed_test_fr}
\end{table*}

\begin{table*}
    \centering
    \begin{tabular}{cc|cccc}
    \hline
translation & aligner & precision & recall & micro-f1 & macro-f1\\
\hline
FAIR & FastAlign & 40.5$_{\pm 2.0}$ & 41.8$_{\pm 1.0}$ & 41.1$_{\pm 1.4}$ & 29.2$_{\pm 1.2}$\\
FAIR & AWESOME w/o co & 73.6$_{\pm 2.7}$ & 71.9$_{\pm 1.1}$ & 72.7$_{\pm 1.6}$ & 55.1$_{\pm 3.9}$\\
FAIR & AWESOME w/ co & 73.6$_{\pm 2.7}$ & 71.9$_{\pm 1.1}$ & 72.7$_{\pm 1.6}$ & 55.1$_{\pm 3.9}$\\
FAIR & AWESOME ft w/o co & 74.7$_{\pm 2.7}$ & 71.1$_{\pm 1.1}$ & 72.8$_{\pm 1.6}$ & 54.9$_{\pm 3.8}$\\
FAIR & AWESOME ft w/ co & \textbf{74.8$_{\pm 2.9}$} & 71.9$_{\pm 1.1}$ & \textbf{73.3$_{\pm 1.7}$} & \textbf{55.4$_{\pm 3.9}$}\\
FAIR & AWESOME pt+ft w/o co & 73.2$_{\pm 2.8}$ & 70.3$_{\pm 1.1}$ & 71.7$_{\pm 1.6}$ & 53.9$_{\pm 3.8}$\\
FAIR & AWESOME pt+ft w/ co & 73.4$_{\pm 2.8}$ & 71.1$_{\pm 1.1}$ & 72.2$_{\pm 1.6}$ & 54.5$_{\pm 3.9}$\\
FAIR ft & FastAlign & 29.7$_{\pm 1.8}$ & 34.3$_{\pm 1.7}$ & 31.8$_{\pm 1.6}$ & 21.7$_{\pm 2.6}$\\
FAIR ft & AWESOME w/o co & 72.6$_{\pm 0.7}$ & 70.8$_{\pm 2.3}$ & 71.6$_{\pm 1.5}$ & 51.9$_{\pm 4.0}$\\
FAIR ft & AWESOME w/ co & 71.3$_{\pm 0.9}$ & 71.9$_{\pm 2.0}$ & 71.6$_{\pm 1.3}$ & 52.4$_{\pm 3.6}$\\
FAIR ft & AWESOME ft w/o co & 73.4$_{\pm 1.0}$ & 71.1$_{\pm 2.5}$ & 72.2$_{\pm 1.7}$ & 52.7$_{\pm 4.5}$\\
FAIR ft & AWESOME ft w/ co & 73.4$_{\pm 1.0}$ & 71.1$_{\pm 2.5}$ & 72.2$_{\pm 1.7}$ & 52.7$_{\pm 4.5}$\\
FAIR ft & AWESOME pt+ft w/o co & 72.9$_{\pm 0.9}$ & \textbf{72.3$_{\pm 2.3}$} & 72.6$_{\pm 1.5}$ & 53.1$_{\pm 3.9}$\\
FAIR ft & AWESOME pt+ft w/ co & 72.9$_{\pm 0.8}$ & 71.8$_{\pm 2.5}$ & 72.3$_{\pm 1.6}$ & 53.0$_{\pm 4.5}$\\
\hline
    \end{tabular}
    \caption{Results of \ttest in German with PubMedBERT.}
    \label{tab:pubmed_test_de}
\end{table*}

\begin{landscape}
\begin{table}[]
    \centering
    \footnotesize
    \begin{tabular}{c|ccc|ccc|ccc|ccc|ccc|ccc|cc}

model & \multicolumn{3}{c|}{Drug} & \multicolumn{3}{c|}{Strength} & \multicolumn{3}{c|}{Frequency} & \multicolumn{3}{c|}{Duration} & \multicolumn{3}{c|}{Dosage} & \multicolumn{3}{c|}{Form} & \multicolumn{2}{c}{global}\\
 & p & r & f1 &p & r & f1 &p & r & f1 &p & r & f1 &p & r & f1 &p & r & f1 & macro & micro\\
\hline
\multicolumn{21}{l}{distilmBERT}\\
\hline
CLT & 52.2 & 82.4 & 63.8 & \textbf{79.1} & \textbf{92.2} & \textbf{85.1} & \textbf{63.1} & \textbf{58.2} & \textbf{60.5} & 87.6 & 88.4 & 88.0 & 67.6 & 67.4 & 67.5 & 55.3 & 38.5 & 44.4 & 68.2 & 65.9 \\
 + realigned & 67.8 & 80.9 & 73.8 & 78.5 & 85.9 & 82.0 & 50.3 & 37.9 & 43.2 & 74.7 & 73.5 & 74.1 & 67.9 & 63.2 & 65.4 & 70.9 & 61.3 & 65.8 & 67.4 & 66.4 \\
Opus - FastAlign & \textbf{69.9} & 77.0 & 73.2 & 76.3 & 84.3 & 80.1 & 48.7 & 44.7 & 46.6 & \textbf{94.2} & \textbf{90.7} & \textbf{92.4} & 64.1 & 70.5 & 67.1 & 67.4 & 62.2 & 64.6 & \textbf{70.7} & \textbf{68.3} \\
Opus - AWESOME pt & 68.1 & 82.7 & 74.7 & 77.1 & 89.0 & 82.5 & 40.5 & 39.2 & 39.8 & 93.8 & \textbf{90.7} & 92.2 & 67.0 & 70.8 & 68.8 & 69.7 & 59.8 & 64.2 & 70.4 & 67.9 \\
Opus - AWESOME ft & 68.0 & \textbf{83.3} & \textbf{74.8} & 74.4 & 87.1 & 80.1 & 33.7 & 33.2 & 33.4 & 93.8 & \textbf{90.7} & 92.2 & 66.6 & 70.5 & 68.5 & 68.2 & 60.0 & 63.7 & 68.8 & 66.2 \\
Opus - AWESOME pt+fr & 65.9 & 82.4 & 73.2 & 76.4 & 89.4 & 82.3 & 39.4 & 38.9 & 39.1 & 92.9 & \textbf{90.7} & 91.8 & \textbf{68.4} & \textbf{72.1} & \textbf{70.2} & 70.1 & 60.4 & 64.8 & 70.2 & 67.8 \\
Opus ft - FastAlign & 68.4 & 80.6 & 74.0 & 74.3 & 85.1 & 79.3 & 45.0 & 44.7 & 44.8 & 91.2 & \textbf{90.7} & 90.9 & 59.7 & 68.2 & 63.6 & 71.0 & \textbf{64.3} & \textbf{67.4} & 70.0 & 67.7 \\
Opus ft - AWESOME pt & 66.3 & 81.8 & 73.2 & 77.0 & 91.8 & 83.7 & 37.1 & 36.3 & 36.7 & 92.9 &\textbf{ 90.7} & 91.8 & 62.4 & 68.4 & 65.3 & 69.8 & 59.1 & 64.0 & 69.1 & 66.5 \\
Opus ft - AWESOME ft & 65.5 & 83.0 & 73.2 & 78.5 & 90.6 & 84.0 & 34.2 & 32.9 & 33.5 & 93.3 & \textbf{90.7} & 92.0 & 61.0 & 69.5 & 65.0 & 70.4 & 59.1 & 64.1 & 68.6 & 65.9 \\
Opus ft - AWESOME pt+fr & 65.0 & 82.7 & 72.8 & 77.6 & 90.2 & 83.4 & 37.3 & 37.9 & 37.6 & 93.3 & \textbf{90.7} & 92.0 & 59.8 & 69.5 & 64.3 & \textbf{71.1} & 58.5 & 64.1 & 69.0 & 66.2 \\
\hline
\multicolumn{21}{l}{XLM-R Base}\\
\hline
CLT & 83.8 & 83.0 & 83.4 & 87.7 & 97.6 & 92.4 & \textbf{73.0} & \textbf{75.3} & \textbf{74.0} & 88.8 & 87.9 & 88.3 & 77.4 & 76.6 & 77.0 & 70.9 & 69.5 & 70.2 & \textbf{80.9} & \textbf{79.1} \\
+ realigned & \textbf{83.9} & 82.8 & \textbf{83.3} & 87.2 & 98.0 & 92.3 & 66.9 & 69.7 & 68.2 & 90.7 & 90.1 & 90.4 & 71.7 & 67.8 & 69.7 & 71.8 & 66.9 & 69.3 & 78.9 & 76.7 \\
Opus - FastAlign & 81.2 & 82.7 & 82.0 & 83.2 & 94.9 & 88.6 & 66.9 & 72.1 & 69.4 & 89.6 & 88.4 & 89.0 & 78.5 & 77.6 & 78.1 & 73.9 & 70.3 & 72.0 & 79.8 & 78.2 \\
Opus - AWESOME pt & 80.5 & 82.4 & 81.4 & 85.3 & 95.3 & 90.0 & 55.3 & 61.6 & 58.1 & \textbf{92.0} & \textbf{90.2} & \textbf{91.1} & 78.3 & \textbf{78.7} & 78.4 & 75.2 & 69.7 & 72.2 & 78.6 & 76.4 \\
Opus - AWESOME ft & 80.5 & 82.4 & 81.4 & 84.9 & 94.9 & 89.6 & 51.0 & 56.1 & 53.4 & 91.5 & 89.8 & 90.6 & 73.1 & 74.2 & 73.6 & \textbf{75.3} & 69.7 & 72.4 & 76.8 & 74.6 \\
Opus - AWESOME pt+fr & 81.1 & 81.8 & 81.4 & 85.1 & 95.7 & 90.1 & 52.1 & 57.9 & 54.8 & 91.4 & 89.3 & 90.4 & \textbf{79.9} & \textbf{78.7} & \textbf{79.3} & 73.2 & 71.2 & 72.2 & 78.0 & 75.8 \\
Opus ft - FastAlign & 83.7 & 82.7 & 83.2 & 84.2 & 96.1 & 89.7 & 70.7 & 73.4 & 72.0 & 90.5 & 87.9 & 89.2 & 75.0 & 75.0 & 75.0 & 72.7 & \textbf{72.3} & \textbf{72.5} & 80.3 & 78.6 \\
Opus ft - AWESOME pt & 81.2 & \textbf{83.3} & 82.2 & 87.3 & \textbf{98.4} & 92.5 & 47.4 & 53.4 & 50.2 & 90.2 & 89.3 & 89.7 & 74.0 & 74.7 & 74.3 & 73.0 & 69.9 & 71.4 & 76.7 & 74.2 \\
Opus ft - AWESOME ft & 81.0 & 82.4 & 81.7 & \textbf{89.0} & 98.0 & \textbf{93.3} & 46.3 & 52.9 & 49.4 & 90.2 & 89.8 & 90.0 & 74.8 & 76.6 & 75.7 & 74.8 & 69.7 & 72.1 & 77.0 & 74.5 \\
Opus ft - AWESOME pt+fr & 80.7 & 82.4 & 81.5 & 87.0 & 97.3 & 91.9 & 54.9 & 61.6 & 58.0 & 89.8 & 89.3 & 89.5 & 78.9 & \textbf{78.7} & 78.8 & 72.5 & 69.9 & 71.2 & 78.5 & 76.3 \\
\hline
\multicolumn{21}{l}{XLM-R Large}\\
\hline
CLT & 80.3 & 81.5 & 80.9 & \textbf{90.5} & 97.3 & \textbf{93.8} & 66.0 & 68.9 & 67.3 & 93.2 & \textbf{89.3} & 91.2 & 76.1 & 75.3 & 75.7 & 75.8 & 67.5 & 71.3 & 80.0 & 77.9 \\
+ realigned & \textbf{85.5} & 82.1 & 83.7 & 90.2 & 97.3 & 93.6 & \textbf{66.6} & 67.6 & 67.1 & 92.7 & 88.8 & 90.7 & \textbf{80.2} & 77.4 & \textbf{78.7} & 74.2 & 68.0 & 70.9 & \textbf{80.8} & \textbf{78.8} \\
Opus - FastAlign & 82.0 & 80.3 & 81.2 & 87.6 & 96.5 & 91.8 & 65.7 & 69.7 & \textbf{67.6} & 92.3 & 88.4 & 90.3 & 76.5 & 76.1 & 76.3 & 78.7 & 66.2 & 71.9 & 79.8 & 78.0 \\
Opus - AWESOME pt & 80.8 & 82.7 & 81.7 & 89.8 & 96.1 & 92.8 & 54.8 & 60.0 & 57.3 & 93.2 & 88.8 & 91.0 & 76.7 & 75.5 & 76.0 & 80.3 & 68.0 & 73.5 & 78.7 & 76.5 \\
Opus - AWESOME ft & 82.7 & 81.2 & 81.9 & 88.1 & 95.7 & 91.7 & 43.9 & 49.5 & 46.5 & 92.7 & 88.4 & 90.5 & 76.8 & 76.8 & 76.8 & \textbf{80.4} & 67.5 & 73.4 & 76.8 & 74.2 \\
Opus - AWESOME pt+fr & 83.8 & 81.2 & 82.4 & 88.8 & 96.5 & 92.5 & 49.0 & 56.6 & 52.5 & 93.2 & 88.8 & 91.0 & 78.3 & \textbf{77.6} & 78.0 & 79.2 & 65.4 & 71.6 & 78.0 & 75.4 \\
Opus ft - FastAlign & 84.6 & 80.6 & 82.5 & 79.0 & 88.2 & 83.3 & 61.5 & \textbf{72.1} & 66.3 & 91.8 & 88.4 & 90.1 & 77.3 & 74.7 & 76.0 & 75.4 & 68.6 & 71.8 & 78.3 & 76.6 \\
Opus ft - AWESOME pt & 83.2 & \textbf{85.1} & \textbf{84.1} & 89.9 & 96.9 & 93.2 & 51.5 & 58.2 & 54.6 & 94.2 & 88.8 & 91.4 & 74.8 & 76.3 & 75.5 & 79.4 & 70.3 & \textbf{74.6} & 78.9 & 76.5 \\
Opus ft - AWESOME ft & 84.4 & 81.8 & 83.0 & 89.9 & \textbf{97.6} & 93.6 & 45.4 & 51.1 & 48.1 & 93.2 & \textbf{89.3} & 91.2 & 74.3 & 76.6 & 75.4 & 75.6 & \textbf{71.4} & 73.4 & 77.4 & 74.8 \\
Opus ft - AWESOME pt+fr & 82.4 & 80.9 & 81.6 & 89.6 & \textbf{97.6} & 93.4 & 51.4 & 56.1 & 53.6 & \textbf{94.6} & \textbf{89.3} & \textbf{91.9} & 74.0 & 76.3 & 75.1 & 75.1 & 70.5 & 72.6 & 78.1 & 75.5 \\
    \end{tabular}
    \caption{Comparison class by class for \ttrain and CLT on \mednerf with multilingual language models.}
    \label{tab:by_class_details}
\end{table}

\end{landscape}

\section{Examples from the different datasets}
\label{app:examples}

    \begin{tabular}
    {>{\raggedright\arraybackslash}p{2cm}%
       >{\raggedright\arraybackslash}p{13cm}%
    }   
        \hline
        \hline
        \multicolumn{2}{c}{\scshape \normalsize  Examples from n2c2}\\
        \hline
        \multicolumn{2}{p{15cm}}{The patient's agitation was managed with \textbf{nightly}$_\text{Frequency}$ \textbf{haldol}$_\text{Drug}$ with \textbf{as needed}$_\text{Frequency}$ \textbf{haldol}$_\text{Drug}$ as well.}\\
        \hline
        \multicolumn{2}{p{15cm}}{Improvement in clinical status was noted overnight and his \textbf{morphine}$_\text{Drug}$ drip was discontinued.}\\
        \hline
        \multicolumn{2}{p{15cm}}{- hold all \textbf{antihypertensives}$_\text{Drug}$ ; plan to add back slowly at reduced doses and varying schedule - rule out MI - \textbf{bolus}$_\text{Dosage}$ \textbf{NS}$_\text{Drug}$ to maintain MAP > 60 with caution given ESRD and oliguric.}\\
        \hline
        \multicolumn{2}{p{15cm}}{\textbf{folic acid}$_\text{Drug}$ \textbf{1 mg}$_\text{Strength}$ \textbf{Tablet}$_\text{Form}$ Sig: \textbf{One (1)}$_\text{Dosage}$ \textbf{Tablet}$_\text{Form}$ PO \textbf{DAILY (Daily)}$_\text{Frequency}$.}\\
        \hline
        \multicolumn{2}{p{15cm}}{\textbf{Iron}$_\text{Drug}$ \textbf{50 mg}$_\text{Strength}$ \textbf{Tablet}$_\text{Form}$ Sustained Release Sig: \textbf{One (1)}$_\text{Dosage}$ \textbf{Tablet Sustained Release}$_\text{Form}$ PO \textbf{once a day}$_\text{Frequency}$.}\\
        \hline
        \hline
        \multicolumn{2}{c}{\vspace{0.1cm}\scshape \normalsize  Examples from GERNERMED test set}\\
        \hline
        \multicolumn{2}{p{15cm}}{Das \textbf{Eplerenon}$_\text{Drug}$ ist wegen Ihrer Herzinsuffizienz. Da können wir jetzt auf \textbf{50 mg}$_\text{Strength}$ p.o. 1-0-0 augmentieren.} \\
        & \textit{Eplerenon is for your heart failure. We can now augment to 50mg p.o. 1-0-0.}\\
        \hline
        \multicolumn{2}{p{15cm}}{Wegen der COPD-Exazerbation wurde \textbf{Terbutalin}$_\text{Drug}$ \textbf{0,25 mg}$_\text{Strength}$ dem Patienten appliziert. Hierfür wurde der subkutane Weg gewählt.}\\
        & \textit{Because of the COPD exacerbation, terbutaline 0.25 mg was administered to the patient. The subcutaneous route was chosen for this.}\\
        \hline
        \multicolumn{2}{p{15cm}}{Bei Vorhofflimmern ist neben \textbf{Betablockern}$_\text{Drug}$ auch die Gabe von \textbf{Magnesium}$_\text{Drug}$ p.o. sinnvoll. Hierfür würden wir mit \textbf{300 mg}$_\text{Strength}$ \textbf{einmal täglich}$_\text{Frequency}$ starten. Sofern möglich, ist eine Einnahme \textbf{mittags}$_\text{Frequency}$ (ca. 12 Uhr) zu bevorzugen.}\\
        &  \textit{In atrial fibrillation, in addition to beta-blocks, the administration of magnesium p.o. is also useful. For this we would start with 300 mg once a day. If possible, it is preferable to take it at noon (around 12 o'clock).}\\
        \hline 
        \multicolumn{2}{p{15cm}}{Zur Optimierung der Herzinsuffizienztherapie wurde die Dosis von \textbf{Sacubitril / Valsartan}$_\text{Drug}$ auf \textbf{97 / 103 mg}$_\text{Strength}$ in \textbf{Tablettenform}$_\text{Form}$ mit Einnahme \textbf{am Morgen und am Abend}$_\text{Frequency}$ erweitert.}\\
        &  \textit{To optimize heart failure therapy, the dose of sacubitril / valsartan was extended to 97~/~103 mg in tablet form with intake in the morning and evening.}\\
        \hline
        \multicolumn{2}{p{15cm}}{Bei bekannter koronarer Herzerkrankung sollte \textbf{lebenslang}$_\text{Duration}$ \textbf{Acetylsalicylsäure}$_\text{Drug}$ \textbf{100 mg}$_\text{Strength}$ \textbf{morgens täglich}$_\text{Frequency}$ in oraler Applikation eingenommen werden.}\\
        &  \textit{In cases of known coronary artery disease, acetylsalicylic acid 100mg should be taken orally daily in the morning as a lifelong treatment.}\\
        \hline
        \hline
        \multicolumn{2}{c}{\vspace{0.1cm}\scshape \normalsize Examples from \mednerf}\\
        \hline
        \multicolumn{2}{p{15cm}}{\textbf{TRAMADOL / PARACETAMOL}$_\text{Drug}$ \textbf{37,5mg / 325mg}$_\text{Strength}$}\\
        &  \textit{TRAMADO/PARACETAMOL 37,5mg/325mg}\\
        \hline
        \multicolumn{2}{p{15cm}}{\textbf{AMLODIPINE}$_\text{Drug}$ \textbf{5 mg}$_\text{Strength}$ ; \textbf{cpr}$_\text{Form}$ \textbf{1}$_\text{Dosage}$ \textbf{comprimé}$_\text{Form}$ \textbf{matin}$_\text{Frequency}$ \textbf{1}$_\text{Dosage}$ \textbf{comprimé}$_\text{Form}$ \textbf{soir}$_\text{Frequency}$}\\
        &  \textit{AMLODIPINE 5mg; tab 1 tablet in the mording 1 tablet in the evening}\\
        \hline
        \multicolumn{2}{p{15cm}}{\textbf{DOLIPRANETABS}$_\text{Drug}$ \textbf{1000 MG}$_\text{Strength}$ \textbf{CPR PELL}$_\text{Form}$ PLQ / 8 (\textbf{Paracétamol}$_\text{Drug}$ \textbf{1.000 mg}$_\text{Strength}$ \textbf{comprimé}$_\text{Form}$)}\\
        &  \textit{DOLIPRANETABS 1000mg TAB PLQ / 8 (Paracetamol 1,000mg tablet)}\\
        \hline
        \multicolumn{2}{p{15cm}}{\textbf{ACIDE ACETYLSALICYLIQUE}$_\text{Drug}$ (\textbf{sel de lysine}$_\text{Drug}$) \textbf{75 mg}$_\text{Strength}$ \textbf{pdre p sol buv sach}$_\text{Form}$ (\textbf{KARDEGIC}$_\text{Drug}$)}\\
        &  \textit{ACETYLSALICYLIC ACID (lysine salts) 75mg oral powder for suspension (KARDEGIC)}\\
        \hline
        \multicolumn{2}{p{15cm}}{\textbf{1}$_\text{Dosage}$ \textbf{sachet}$_\text{Form}$ \textbf{matin midi et soir}$_\text{Frequency}$ si besoin}\\
        &  \textit{1 packet in the morning, at noon, and in the evening, if needed}\\
        \hline
    \end{tabular}

\end{document}